%% file: draft.tex
\documentclass[10pt,twocolumn,letterpaper]{article}
\usepackage[svgnames,table]{xcolor}

\usepackage{epsfig}
\usepackage{graphicx}
\usepackage{amsmath}
\usepackage{amssymb}
\usepackage{listings}
\usepackage{makecell}
\usepackage{booktabs}
\usepackage{multirow}
\usepackage{color}
\usepackage{comment}
\usepackage{courier}

\usepackage{icomma}
\usepackage{enumitem}
\usepackage{amsthm, url}

\usepackage[bf,font=footnotesize]{caption}

\usepackage{subfig}
\usepackage{xspace}

\makeatletter
\DeclareRobustCommand\onedot{\futurelet\@let@token\@onedot}
\def\@onedot{\ifx\@let@token.\else.\null\fi\xspace}

\def\eg{\emph{e.g}\onedot} 
\def\ie{\emph{i.e}\onedot}

\def\etal{\emph{et al}\onedot}
\makeatother

\def\sexyname{{Ahab}\xspace}
\def\KBKnowl{{KB-knowledge}\xspace} %
\def\KBName{{KB-VQA}\xspace}
\begin{document}

\title{Explicit Knowledge-based Reasoning for Visual Question Answering}

\author{  Peng Wang$^*$, Qi Wu\footnote{The first two authors contributed to this work equally.}, Chunhua Shen, Anton van den Hengel, Anthony Dick \\
School of Computer Science, The University of Adelaide
}

\maketitle
\begin{abstract}

We describe a method for visual question answering which is capable of reasoning about contents of an image on the basis of information extracted from a large-scale knowledge base.  The method not only answers natural language questions using concepts not contained in the image, but can provide an explanation of the reasoning by which it developed its answer.
The method is capable of answering far more complex questions than the predominant long short-term memory-based approach, and outperforms it significantly in the testing.
We also provide a dataset and a protocol by which to evaluate such methods, thus addressing one of the key issues in general visual question answering.

\end{abstract}

\input{intro.tex}

\input{relatedwork.tex}
\input{dataset.tex}
\input{baseline.tex}

\input{exp.tex}

\vspace{-10pt}
\section{Conclusion}
\vspace{-10pt}
\label{Sec:Conclusion}

We have described a method capable of reasoning about the content of general images, and interactively answering a wide variety of questions about them.  The method develops a structured representation of the content of the image, and relevant information about the rest of the world, on the basis of a large external knowledge base.  It is capable of explaining its reasoning in terms of the entities in the knowledge base, and the connections between them.
\sexyname is applicable to any knowledge base for which a SPARQL interface is available.  This includes any of the over a thousand RDF datasets online~\cite{schmachtenberg2014state} which relate information on taxonomy, music, UK government statistics, Brazilian politicians, and the articles of the New York Times, amongst a host of other topics.  Each could be used to provide a specific visual question answering capability, but many can also be linked by common identifiers to form larger repositories.  If a knowledge base containing common sense were available, the method we have described could use it to draw sensible general conclusions about the content of images.

We have also provided a dataset and methodology for testing the performance of general visual question answering techniques, and shown that \sexyname substantially outperforms the currently predominant visual question answering approach when so tested.

{\small
\bibliographystyle{ieee}
\bibliography{CSRef}
}
\clearpage
\appendix
\input{appendix.tex}

\end{document}

%% file: intro.tex
\vspace{-5pt}
\section{Introduction}

Visual Question Answering (VQA) requires that a method be able to interactively answer questions about images.  The questions are typically posed in natural language, as are the answers.  The problem requires image understanding, natural language processing, and a means by which to relate images and text.   More importantly, however, the interactivity implied by the problem means that it cannot be determined beforehand which questions will be asked.  This requirement to answer a wide range of image-based questions, on the fly, means that the problem is closely related to some of the ongoing challenges in Artificial Intelligence~\cite{geman2015visual}.

\begin{figure}[t!]
\begin{center}
   \includegraphics[width=1.0\linewidth]{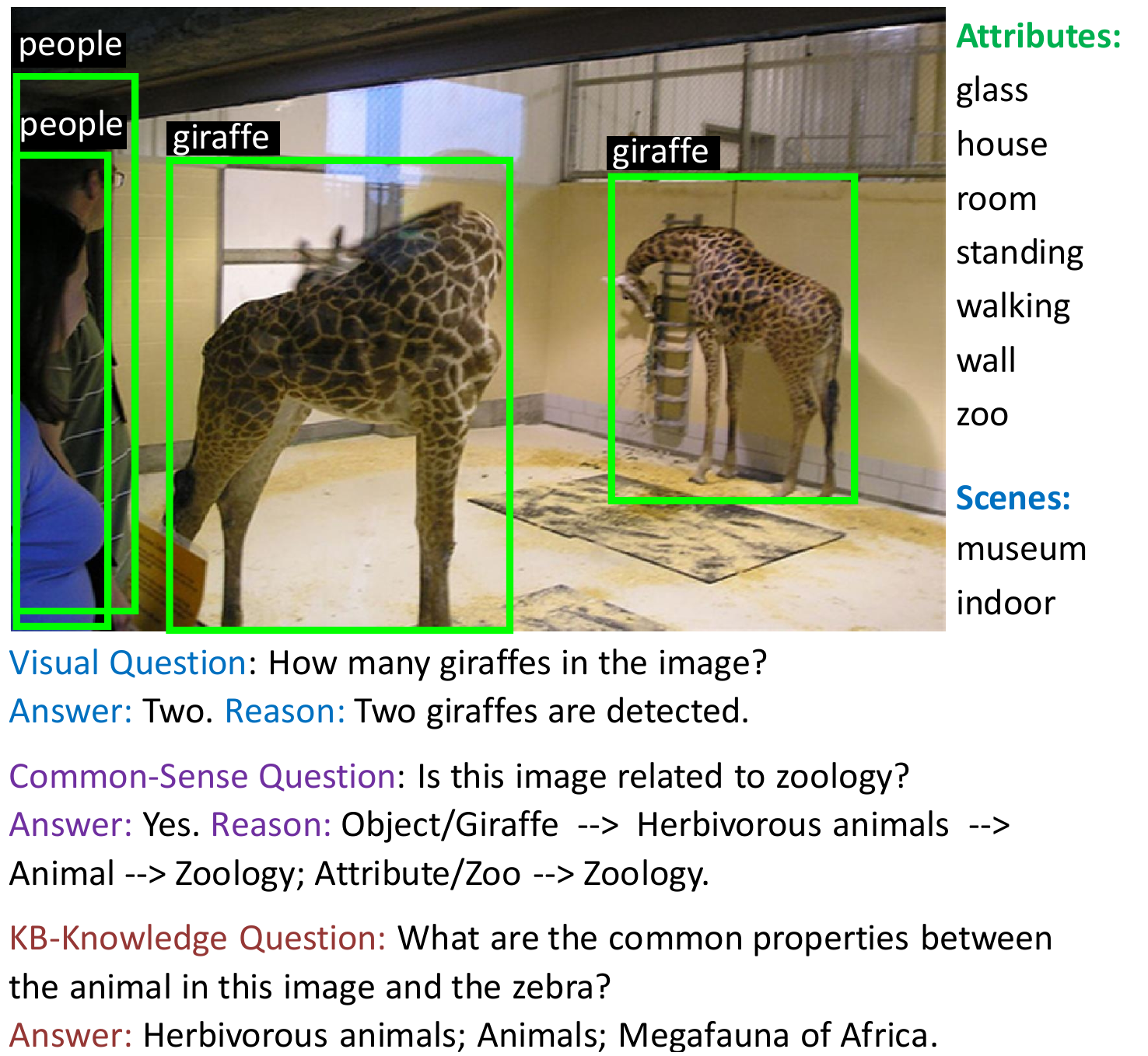}
\end{center}
\vspace{-15pt}
   \caption{A real example of the proposed \KBName dataset and the results given by \sexyname, the proposed VQA approach.
            The questions in the collected dataset are separated into three classes with different knowledge levels:  ``Visual'', ``Common-sense'' and ``\KBKnowl''.
            Our approach answers questions by extracting several types of visual concepts (object, attribute and scene class) 
            from an image and aligning them to large-scale structured knowledge bases. 
            Apart from answers, our approach can also provide reasons/explanations for certain types of questions.}
\vspace{-20pt}
\label{fig:real_example}
\end{figure}

Despite the implied need to perform general reasoning about the content of images, most VQA methods perform no explicit reasoning at all 
(see, for example,~\cite{antol2015vqa,gao2015you,malinowski2015ask,ren2015image}).  
The predominant method~\cite{gao2015you,malinowski2015ask} is based on forming a direct connection between a convolutional neural network (CNN)~\cite{krizhevsky2012imagenet,simonyan2014very,szegedy2014going} to perform the image analysis, and a Long Short-Term Memory (LSTM)~\cite{hochreiter1997long} network to process the text.  This approach performs very well in answering simple questions directly related to the content of the image, such as `What color is the ...?', or `How many ... are there?'.%

There are a number of problems with the LSTM approach, however.  The first is that the method does not explain how it arrived at its answer.  This means that it is impossible to tell whether it is answering the question based on image information, or just the prevalence of a particular answer in the training set.  The second problem is that the amount of prior information that can be encoded within a LSTM system is very limited.  
DBpedia~\cite{auer2007dbpedia}, with millions of concepts and hundred millions of relationships, 
contains a small subset of the information required to truly reason about the world in general.  Recording this level of information would require an implausibly large LSTM, and the amount of training data necessary would be completely impractical.  The third, and major, problem with the LSTM approach is that it is incapable of explicit reasoning except in very limited situations~\cite{rocktaschel2015reasoning}.  

We thus propose \sexyname\footnote{Ahab, the captain in the novel Moby Dick, is either a brilliant visionary, or a deluded fanatic, depending on your perspective.}, a new approach to VQA which is based on explicit reasoning about the content of images.  \sexyname first detects relevant content in the image, and relates it to information available in a knowledge base.  A natural language question is processed into a suitable query which is run over the combined image and knowledge base information.  This query may require multiple reasoning steps to satisfy.  The response to the query is then processed so as to form the final answer to the question.

This process allows complex questions to be asked which rely on information not available in the image.  Examples include questioning the relationships between two images, or asking whether two depicted animals are close taxonomic relatives.

\subsection{Background}
\label{sec:Background}

The first VQA approach~\cite{malinowski2014multi} proposed to process questions using semantic parsing~\cite{liang2013learning} and obtain answers through Bayesian reasoning.
Both~\cite{gao2015you} and~\cite{malinowski2015ask} used CNNs to extract image features 
and relied on LSTMs to encode questions and decode answers.
The primary distinction between these approaches is that~\cite{gao2015you} used independent LSTM networks for question encoding and answer decoding, while~\cite{malinowski2015ask} used one LSTM for both tasks.  
Irrespective of the finer details, we label this the LSTM approach.

Datasets are critical to VQA, as the generality of the questions asked, and the amount of training data available,  have a large impact on the set of methods applicable. Malinowski and Fritz~\cite{malinowski2014towards} proposed the DAQUAR dataset, which is mostly composed of questions requiring only visual knowledge.
Ren \etal~\cite{ren2015image} constructed a VQA dataset (refer to as TORONTO-QA), 
which generated questions automatically by transforming captions on MS COCO~\cite{lin2014microsoft} images.
The COCO-VQA dataset~\cite{antol2015vqa} is currently the largest VQA dataset, which contains $300$K questions and answers about MS COCO images.

Answering general questions posed by humans about images inevitably requires reference to information not contained in the image itself.  To an extent this information may be provided by an existing training set such as ImageNet~\cite{deng2009imagenet}, or MS COCO~\cite{lin2014microsoft} as class labels or image captions.  This approach is inflexible and does not scale, however, and cannot provide the wealth of background information required to answer even relatively simple questions about images.
This has manifested itself in the fact that it has proven very difficult to generate a set of image-based questions which are simple enough that VQA approaches can actually 
answer them~\cite{antol2015vqa,malinowski2014towards,ren2015image}.

Significant advances have been made, however, in the construction of large-scale {\em structured} Knowledge Bases 
(KBs)~\cite{auer2007dbpedia,banko2007open,bollacker2008freebase,carlson2010toward,chen2013neil,mahdisoltani2014yago3,vrandevcic2014wikidata}. 
In structured KBs,  knowledge is typically represented by a large number of triples of the form \texttt{(arg1,rel,arg2)}, 
where \texttt{arg1} and \texttt{arg2} denote two entities in the KB and \texttt{rel} denotes a predicate representing the relationship between these two entities.  A collection of such triples can be seen as a large interlinked graph.
Such triples are often described in terms of a Resource Description Framework~\cite{rdf2014resource} (RDF) specification, and housed in a  relational database management system (RDBMS), or triple-store, which allows queries over the data.
The knowledge that ``a cat is a domesticated animal'', for instance, is stored in an RDF KB by the triple \texttt{(cat,is-a,domesticated animal)}.
The information in KBs can be accessed efficiently using a query language.  
In this work we use SPARQL Protocol~\cite{prud2008sparql} to query the OpenLink Virtuoso~\cite{erling2012virtuoso} RDBMS.  
For example, the query \texttt{?x:(?x,is-a,domesticated animal)} returns all domesticated animals in the graph.

Popular large-scale structured KBs are constructed
either by manual-annotation/crowd-sourcing
(\eg, DBpedia~\cite{auer2007dbpedia}, Freebase~\cite{bollacker2008freebase} and Wikidata~\cite{vrandevcic2014wikidata}),
or by automatically extracting from unstructured/semi-structured data 
(\eg, YAGO~\cite{hoffart2013yago2,mahdisoltani2014yago3}, OpenIE~\cite{banko2007open,etzioni2011open,fader2011identifying}, 
NELL~\cite{carlson2010toward}, NEIL~\cite{chen2013neil,chen2014enriching}). 
The KB we use here is DBpedia,
which contains structured information extracted from Wikipedia.
Compared to KBs extracted automatically from unstructured data (such as OpenIE), the data in DBpedia is more accurate and has a well-defined ontology.  The method we propose is applicable to any KB that admits SPARQL queries, however, including those listed above and the huge variety of subject-specific RDF databases available.

The advances in structured KBs, have driven an increasing interest in the NLP and AI communities
in the problem of natural language question answering using structured KBs 
(refer to as KB-QA)~\cite{berant2013semantic,bordes2014question,cai2013large,fader2014open,kolomiyets2011survey,kwiatkowski2013scaling,yao2014information,liang2013learning,unger2012template}. 
The VQA approach which is closest to KB-QA (and to our approach) is that of Zhu \etal~\cite{zhu2015building}
as they use a KB and RDBMS to answer image-based questions.
They build the KB for the purpose, however, using an MRF model, with image features and scene/attribute/affordance labels as nodes. The undirected links between nodes represent mutual compatibility/incompatibility relationships. The KB thus relates specific images to specified image-based quantities to the point where the database schema prohibits recording general information about the world.
The queries that this approach can field are crafted in terms of this particular KB, and thus relate to the small number of attributes specified by the schema.  The questions are framed in an RDMBS query language, rather than natural language.

\subsection{Contribution}

Our primary contribution is a method we label \sexyname for answering a wide variety of questions about images that require external information to answer.  The method accepts questions in natural language, and answers in the same form.  It is capable of correctly answering a far broader range of image-based questions than competing methods, and provides an explanation of the reasoning by which it arrived at the answer.  \sexyname exploits DBpedia as its source of external information, and requires no VQA training data (it does use ImageNet and MS COCO to train the visual concept detector).

We also propose a dataset, and protocol for measuring performance, for general visual question answering.
The questions in the dataset are generated by human subjects based on a number of pre-defined templates.
The questions are given one of three labels reflecting the information required to answer them: ``Visual'', ``Common-sense'' and ``\KBKnowl'' (see Fig.~\ref{fig:real_example}).
Compared to other VQA datasets~\cite{antol2015vqa,lin2014microsoft,malinowski2014towards,ren2015image}, the questions in the \KBName dataset, as a whole, require a higher level of external knowledge to answer.
It is expected that humans will require  the help of 
Wikipedia to answer ``\KBKnowl'' questions.
The evaluation protocol requires human evaluation of question answers, as this is the only practical method of testing which does not place undue limits on the questions which can be asked.

%% file: dataset.tex
\section{The \KBName Dataset}
\label{sec:dataset}

The \KBName dataset has been constructed for the purpose of evaluating the performance of VQA algorithms capable of answering
higher knowledge level questions and explicit reasoning about image contents using external information.

\begin{table*}[t!]
\centering
\small
\begin{tabular}{lll}
\hline
Name & Template & Num. \\ \hline
{\em IsThereAny} 	& Is there any {$\langle${\em concept}$\rangle$}? & $419$ \\
{\em IsImgRelate} 	& Is the image related to {$\langle${\em concept}$\rangle$}? & $381$ \\
{\em WhatIs} 		& What is the {$\langle${\em obj}$\rangle$}? & $275$  \\
{\em ImgScene} 		& What scene does this image describe? & $263$ \\
{\em ColorOf} 		& What color is the {$\langle${\em obj}$\rangle$}? & $205$ \\
{\em HowMany} 		& How many {$\langle${\em concept}$\rangle$} in this image? & $157$ \\
{\em ObjAction} 	& What is the {$\langle${\em person/animal}$\rangle$} doing? & $147$ \\
{\em IsSameThing}	& Are the {$\langle${\em obj1}$\rangle$} and the {$\langle${\em obj2}$\rangle$} the same thing? & $71$ \\
{\em MostRelObj} 	& Which {$\langle${\em obj}$\rangle$} is most related to {$\langle${\em concept}$\rangle$}? & $56$ \\
{\em ListObj} 		& List objects found in this image. & $54$ \\
{\em IsTheA} 		& Is the {$\langle${\em obj}$\rangle$} a {$\langle${\em concept}$\rangle$}? & $51$ \\
{\em SportEquip} 	& List all equipment I might use to play this sport. & $48$ \\
{\em AnimalClass} 	& What is the {$\langle${\em taxonomy}$\rangle$} of the {$\langle${\em animal}$\rangle$}? & $46$ \\
{\em LocIntro} 		& Where was the {$\langle${\em obj}$\rangle$} invented? & $40$ \\
{\em YearIntro} 	& When was the {$\langle${\em obj}$\rangle$} introduced? & $32$ \\
{\em FoodIngredient} 	& List the ingredient of the $\langle \mathrm{food} \rangle$. & $31$ \\
{\em LargestObj} 	& What is the largest/smallest {$\langle${\em concept}$\rangle$}? & $27$ \\
{\em AreAllThe} 	& Are all the {$\langle${\em obj}$\rangle$} {$\langle${\em concept}$\rangle$}? & $27$ \\
{\em CommProp} 		& List the common properties of the {$\langle${\em obj1}$\rangle$} and {$\langle${\em concept/obj2}$\rangle$}. & $26$ \\
{\em AnimalRelative}	& List the close relatives of the {$\langle${\em animal}$\rangle$}. & $17$ \\
{\em AnimalSame} 	& Are {$\langle${\em animal1}$\rangle$} and {$\langle${\em animal2}$\rangle$} in the same {$\langle${\em taxonomy}$\rangle$}? & $17$ \\
{\em FirstIntro} 	& Which object was introduced earlier, {$\langle${\em obj1}$\rangle$} or {$\langle${\em concept/obj2}$\rangle$}? & $8$ \\
{\em ListSameYear} 	& List things introduced in the same year as the {$\langle${\em obj}$\rangle$}. & $4$ \\
 \hline
\end{tabular} %
\caption{Question templates in descending order of number of instantiations. The total number of questions is 2402.
Note that some templates allow questioners to ask the same question in different forms.}
\label{tab:templates}
\end{table*}

\subsection{Data Collection}

\noindent\textbf{Images}
We select $700$ of the validation images from the MS COCO~\cite{lin2014microsoft} dataset due to the rich contextual information and diverse object classes therein.
The images are selected so as to cover around $150$ object classes and $100$ scene classes, and typically exhibit $6$ to $7$ objects each.

\noindent\textbf{Templates}
Five human subjects (questioners) are asked to generate $3$ to $5$ question/answer pairs for each of the $700$ images,
by instantiating the $23$ templates shown in Table~\ref{tab:templates}.
There are several slots to be filled in these templates:

\noindent $-$
{$\langle${\em obj}$\rangle$}
is used to specify the visual objects in an image, which can be a single word ``object'' or its super-class name like ``animal'', ``food'' or ``vehicle''.
We also allow questioners to specify a visual object using its size (\eg, small, large)
or location (\eg, left, right, top, bottom, center).

\noindent $-$
{$\langle${\em person}$\rangle$}/{$\langle${\em animal}$\rangle$}/{$\langle${\em food}$\rangle$} is used to specify a person/animal/food
with optionally size or location.

\noindent $-$
{$\langle${\em concept}$\rangle$} can be filled by any word or phrase
which probably corresponds to an entity in DBpedia.

\noindent $-$
{$\langle${\em taxonomy}$\rangle$} corresponds to the taxonomy of animals, including kingdom, phylum, class, order, family or genus.

\noindent\textbf{Questions}
The questions of primary interest here are those require knowledge external to the image to answer.
Each question has a label reflecting the human-estimated level of knowledge required to answer it correctly.
The ``Visual'' questions can be answered directly using visual concepts gleaned from ImageNet and MS COCO (such as ``Is there a dog in this image?'');
``Common-sense'' questions should not require an adult to refer to an external source (``How many road vehicles in this image?'');
while answering  ``\KBKnowl'' questions is expected to require Wikipedia or similar (``When was the home appliance in this image invented?'').

\subsection{Data Analysis}

\begin{figure}[tbp!]
\begin{center}
   \includegraphics[width=0.977\linewidth]{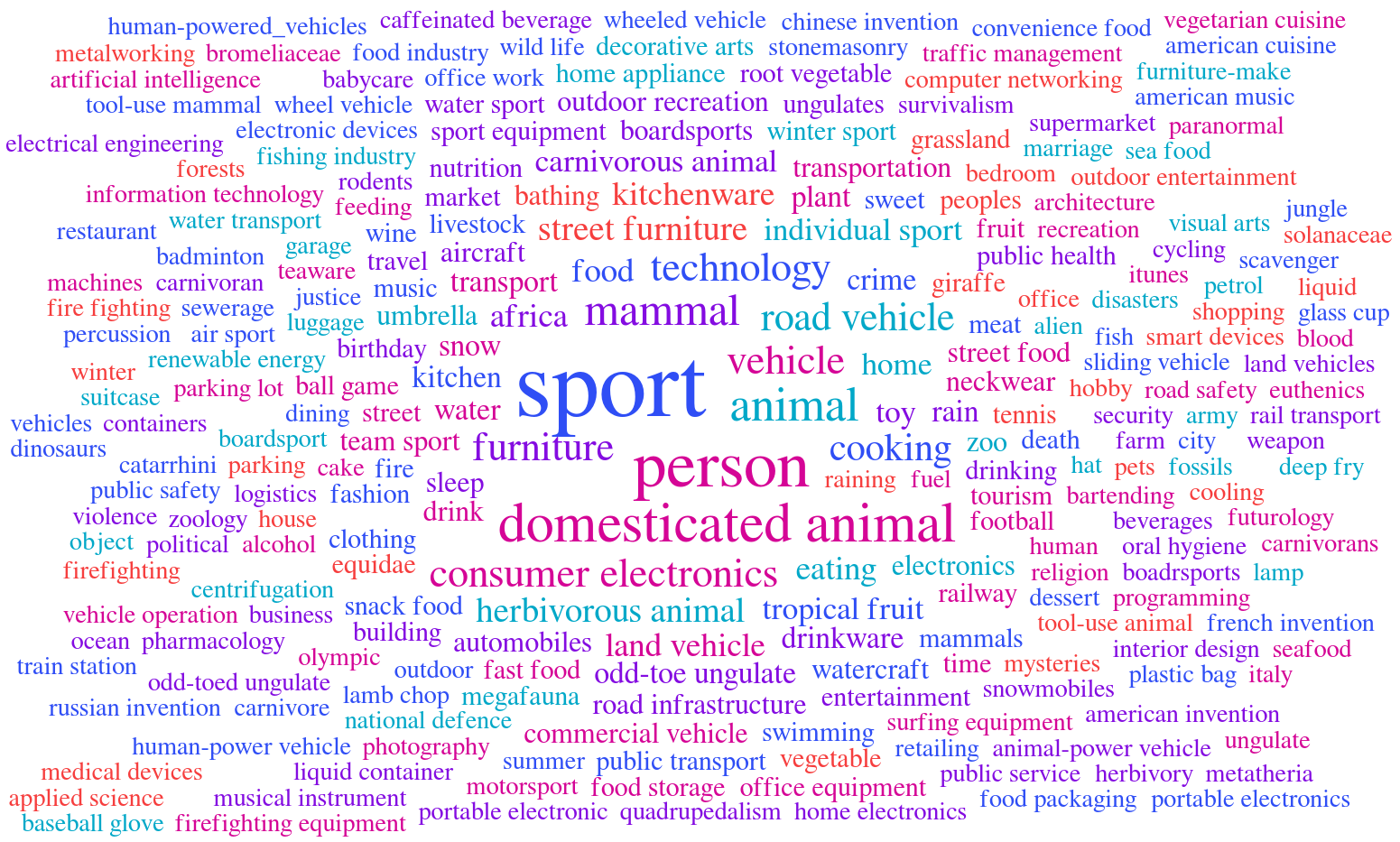}
\end{center}
\vspace{-15pt}
   \caption{The cloud of $254$ {$\langle${\em concept}$\rangle$}-phrases for questions in levels ``Common-sense'' and ``\KBKnowl'', with size representing the frequency.
            $55\%$ of these phrases have less than $20$ mentions in the $300$K questions of the COCO-VQA dataset.
            }
\vspace{-15pt}
\label{fig:concepts}
\end{figure}

\begin{figure*}
\centering
\includegraphics[width=0.97\linewidth]{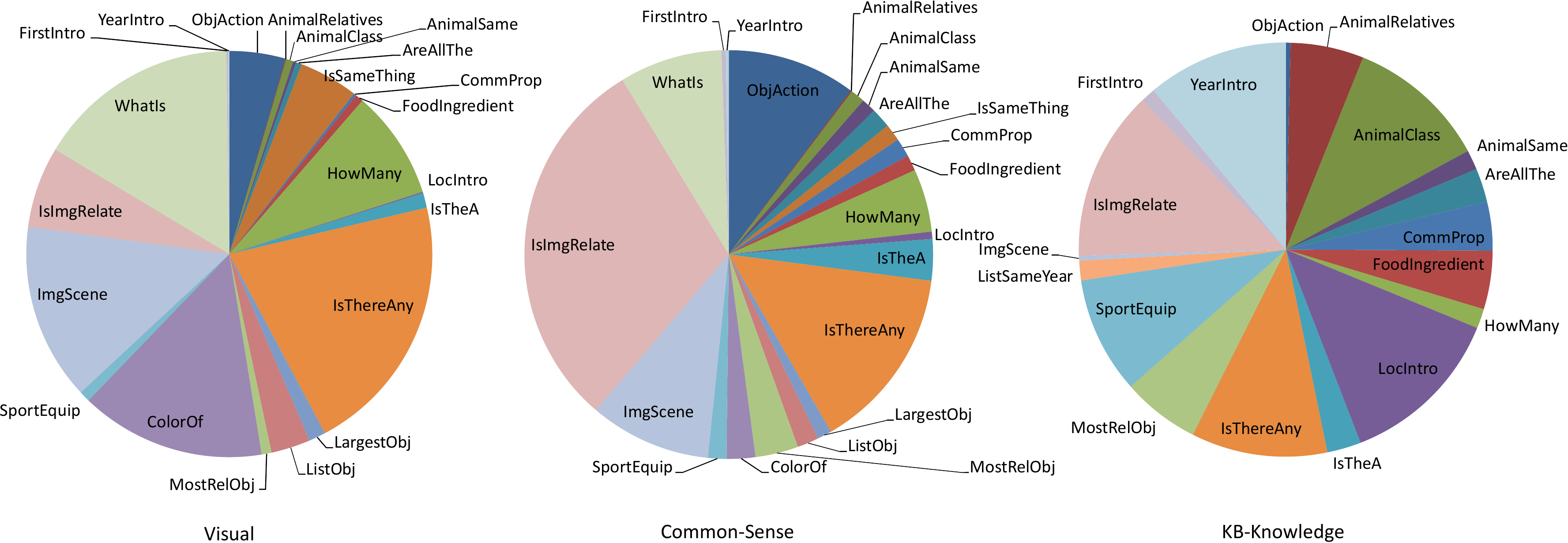}
\vspace{-8pt}
\caption{Question template frequencies for different knowledge levels.
}
\label{fig:questiontypes}
\vspace{-12pt}
\end{figure*}

\noindent\textbf{Questions}
From Table~\ref{tab:templates}, we see that the top $5$ most frequently used templates are {\em IsThereAny}, {\em IsImgRelate}, {\em WhatIs}, {\em ImgScene} and {\em ColorOf}.
Some templates lead to questions that can be answered without any external knowledge, such as {\em ImgScene} and {\em ListFound}.
But ``Common-sense'' or ``\KBKnowl'' is required to analyze the relationship between visual objects and concepts in questions like {\em IsTheA}, {\em AreAllThe} and {\em IsThereAny}.
More complex  questions like {\em YearInvent} and {\em AniamlClass} demand ``\KBKnowl''.
The total number of questions labelled ``Visual'', ``Common-sense'' and ``\KBKnowl'' are $1256$, $883$ and $263$ respectively.  So answering around half of the questions will require external knowledge.
Fig.~\ref{fig:questiontypes} shows the distribution of the $23$ templates for each question type.
Templates {\em IsImgRelate} and {\em IsThereAny} cover almost half of the ``Common-sense'' questions.
There are $18$ templates shown for ``\KBKnowl'' questions (as {\em WhatIs}, {\em IsSameThing}, {\em ListObj}, {\em LargestObj} and {\em ColorOf}  do not appear),
which exhibit a more balanced distribution.

In total, $330$ different phrases were used by questioners in filling the {$\langle${\em concept}$\rangle$} slot.
There were $254$ phrases (see Fig.~\ref{fig:concepts}) used in questions requiring external knowledge (\ie, ``Common-sense'' and ``\KBKnowl''),
$55\%$ of which are mentioned very rarely (less than $20$ times) in the $300$K questions of the COCO-VQA~\cite{antol2015vqa} dataset.
For the ``\KBKnowl'' level, $67$ phrases are used and greater than $85\%$ of them have less than $20$ mentions in COCO-VQA.
Examples of concepts not occurring in COCO-VQA include
``logistics'', ``herbivorous animal'', ``animal-powered vehicle'', ``road infrastructure'' and  ``portable electronics''.

Compared to other VQA datasets, a large proportion of the questions in \KBName require external knowledge to answer.
The questions defined in DAQUAR~\cite{malinowski2014towards} are almost exclusively ``Visual'' questions, referring to ``color'', ``number'' and ``physical location of the object''.
In the TORONTO-QA dataset~\cite{ren2015image}, questions are generated automatically from image captions which describe the major visible content of the image.
For the COCO-VQA dataset~\cite{antol2015vqa}, only $5.5\%$ of questions require adult-level common-sense,
and none requires ``\KBKnowl'' (by observation).

\noindent\textbf{Answers}
Questions starting with ``Is ...'' and ``Are ...'' require logical answers (
$61\%$ ``yes'' and $39\%$ ``no'').
Questions starting with ``How many'', ``What color'', ``Where'' and ``When'' need to be answered with number, color, location and time respectively.
Of the human answers for ``How many'' questions,  $74\%$ are less than or equal to $5$.
The most frequent number answers are ``$1$''($58$ occurrences), ``$2$''($66$) and ``$3$''($44$).
We also have $16$ ``How many'' questions with human answer ``$0$''.
The answers for ``What ...'' and ``List ...'' vary significantly, covering a wide range of concepts.

%% file: baseline.tex
\section{The \sexyname VQA approach}
\label{sec:baseline}

\subsection{RDF Graph Construction}

In order to reason about the content of an image we need to amass the relevant information.  This is achieved by detecting concepts in the query image and linking them to the relevant parts of the KB.

\noindent\textbf{Visual Concepts}
Three types of visual concepts are detected in the query image, including:
\noindent {\em $-$ Objects:} 
We trained two Fast-RCNN~\cite{girshick2015fast} detectors on 
MS COCO $80$-class objects~\cite{lin2014microsoft} and ImageNet $200$-class objects~\cite{deng2009imagenet}.
Some classes with low precision were removed from the models, such as ``ping-pong ball'' and ``nail''.
The finally merged detector contains $224$ object classes, which can be found in the supplementary material.

\noindent {\em $-$ Image Scenes:} 
The scene classifier is obtained from \cite{zhou2014learning}, which is a VGG-$16$~\cite{simonyan2014very} CNN model trained on the MIT Places$205$ dataset~\cite{zhou2014learning}.
In our system, the scene classes corresponding to the top-$3$ scores are selected.

\noindent {\em $-$ Image Attributes:}
In the work of \cite{qi2015caption}, a VGG-$16$ CNN pre-trained on ImageNet is fine-tuned on the MS COCO image-attribute training data.
The vocabulary of attributes defined in \cite{qi2015caption} covers a variety of high-level concepts related to an image, such as actions, objects, sports and scenes. 
We select the top-$10$ attributes for each image.

\begin{figure}[tbp!]
\begin{center}
   \includegraphics[width=0.9\linewidth]{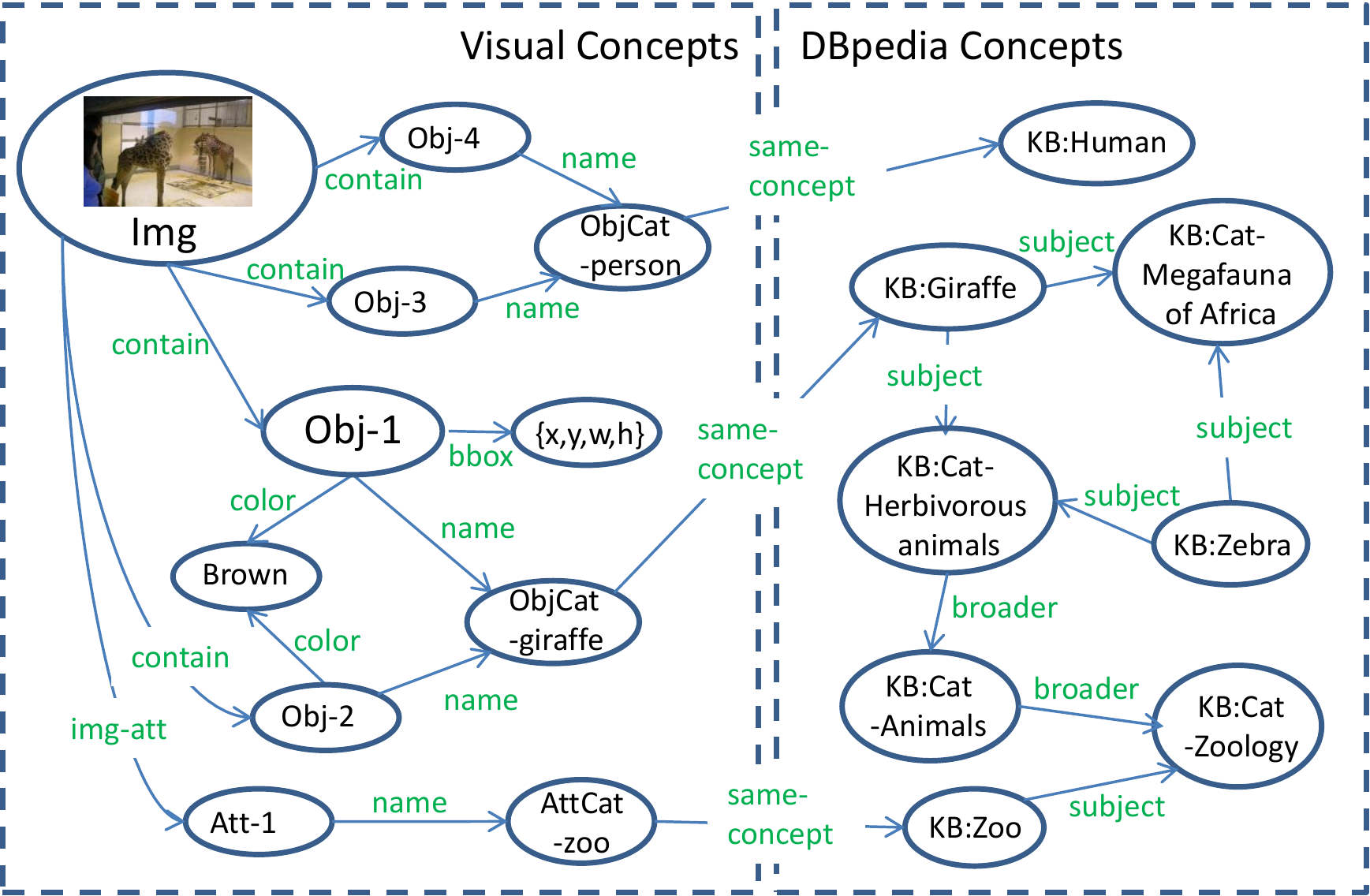}
\end{center}
\vspace{-15pt}
   \caption{ A visualisation of an RDF graph such as might be constructed by \sexyname. 
             The illustrated entities are relevant to answering the questions in Fig.~\ref{fig:real_example}
             (many others not shown for simplicity).   
             Each arrow corresponds to one triple in the graph, with circles representing entities and green text reflecting predicate type.
             The graph of extracted visual concepts (left side) is linked to DBpedia (right side) 
             by mapping object/attribute/scene categories to DBpedia entities using the predicate {\ttfamily same-concept}. }
\vspace{-15pt}
\label{fig:linked_graph}
\end{figure}

\noindent\textbf{Linking to the KB}
Having extracted a set of concepts of interest from the image, we now need to relate them to the appropriate information in the KB.
As shown in the left side of Fig.~\ref{fig:linked_graph}, the visual concepts (object, scene and attribute categories) are stored as RDF triples.
For example, the information that ``The image contains a giraffe object'' is expressed as:
{\ttfamily (Img,}{\ttfamily contain,}{\ttfamily Obj-1)} and {\ttfamily (Obj-1,}{\ttfamily name,}{\ttfamily ObjCat-giraffe)}.
Each visual concept is linked to DBpedia entities with the same semantic meaning (identified through a uniform resource identifier\footnote{In DBpedia, 
each entity has a uniform resource identifier (URI). For example, the animal giraffe corresponds to URI: \url{http://dbpedia.org/resource/Giraffe}.} (URI)), for example {\ttfamily (ObjCat-giraffe,} {\ttfamily same-concept,} {\ttfamily KB:Giraffe)}.
The resulting RDF graph includes all of the relevant information in DBpedia, linked as appropriate to the visual concepts extracted from the query image. 
This combined image and DBpedia information is then accessed through a local OpenLink Virtuoso~\cite{erling2012virtuoso} RDBMS.

\subsection{Answering Questions}
\label{sec:answering_engines}
Having gathered all of the relevant information from the image and DBpedia, we now use them to answer questions.

\begin{figure}[tbp!]
\begin{center}
   \includegraphics[width=0.9\linewidth]{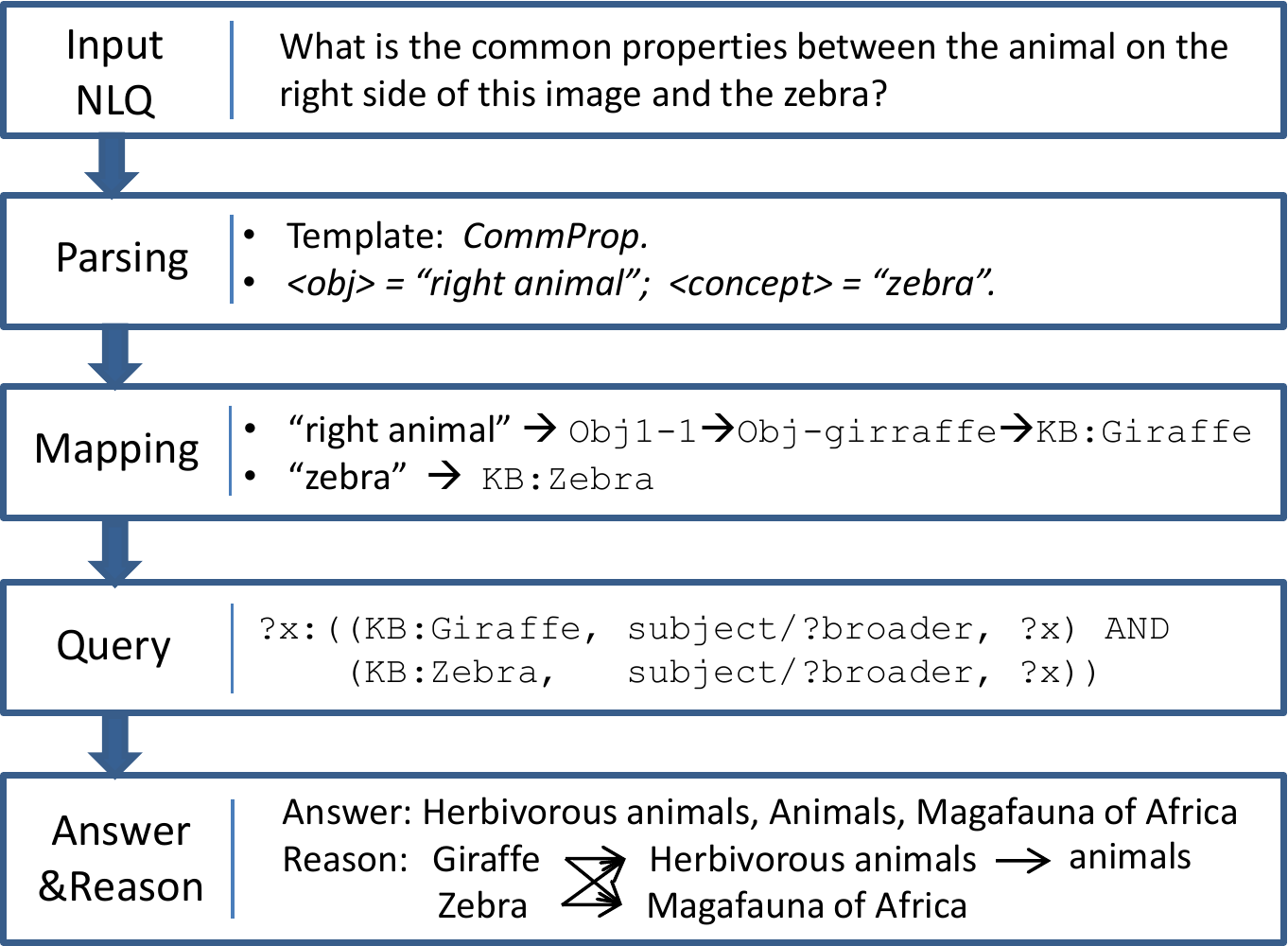}
\end{center}
\vspace{-15pt}
   \caption{The question processing pipeline. The input question is parsed using a set of NLP tools, and the appropriate template identified. 
            The extracted slot-phrases are then mapped to entities in the KB. 
            Next, KB queries are generated to mine the relevant relationships for the KB-entities. 
            Finally, the answer and reason are generated based on the query results.
            The predicate {\ttfamily category/?broader} is used to obtain the categories transitively (see \cite{prud2008sparql} for details). }
\vspace{-15pt}
\label{fig:flow_chart}
\end{figure}

\noindent\textbf{Parsing NLQs}
Given a question posed in natural language, we first need to translate it to a format which can be used to query the RDBMS.
Quepy\footnote{\url{http://quepy.readthedocs.org/en/latest/}} is a Python framework  designed within the NLP community to achieve exactly this task.
To achieve this Quepy requires a set of templates, framed in terms of regular expressions. It is these templates which form the basis of Table~\ref{tab:templates}.
Quepy begins by tagging each word in the question using NLTK~\cite{bird2009natural}, which is composed of a tokenizer, a part-of-speech tagger and a lemmatizer. 
The tagged question is then parsed by a set of regular expressions (regex), each defined for a specific question template.
These regular expressions are built using REfO\footnote{\url{https://github.com/machinalis/refo}} to increase the flexibility of question expression as much as possible.
Once a regex matches the question, it will extract the slot-phrases and forward them 
for further processing.
For example, the question in Fig.~\ref{fig:flow_chart} is matched to template {\em CommProp} and 
the slot-phrases for $\langle${\em obj}$\rangle$ and $\langle${\em concept}$\rangle$ are ``right animal'' and ``zebra'' respectively.

\noindent\textbf{Mapping Slot-Phrases to KB-entities}
Note that the slot-phrases are still expressed in natural language. 
The next step is to find the correct correspondences
between the slot-phrases and entities in the constructed graph.

\noindent $-$
Slots {$\langle${\em obj}$\rangle$}/{$\langle${\em animal}$\rangle$}/{$\langle${\em food}$\rangle$}
correspond to objects detected in the image, and are identified by comparing provided locations, sizes and names with information in the RDF graph recovered from the image (bounding boxes, for example).
This process is heuristic, and forms part of the Quepy rules.
In Fig.~\ref{fig:flow_chart}, ``right animal'' is thus mapped to the entity {\ttfamily Obj-1} in the linked graph (see Fig.~\ref{fig:linked_graph}).

\noindent $-$
Phrases in slot {$\langle${\em concept}$\rangle$} need to be mapped to entities in DBpedia.
In \sexyname this mapping is conducted by string matching between phrases and entity names.
The DBpedia predicate {\ttfamily wikiPageRedirects} is used to handle synonyms, capitalizations, punctuation, tenses, abbreviation and misspellings (see the supplementary material for details).
In Fig.~\ref{fig:flow_chart}, we can see that the phrase ``zebra'' is mapped to the KB-entity {\ttfamily KB:Zebra}.

\noindent\textbf{Query Generation}
With all concepts mapped to KB entities, the next step is to 
form the appropriate SPARQL queries, depending on the question template.
The following several types of DBpedia predicates are used extensively for generating queries and further analysis:

\noindent $-$ {\em Infoboxes} in Wikipedia provide a variety of relationships for different types of entities,
such as animal taxonomy and food ingredient.

\noindent $-$ {\em Wikilinks} are extracted from the internal links between Wikipedia articles.
Compared to Infoboxes, Wikilinks are generic and less precise as the relation property is unknown.
There are $162$ million such links in the DBpedia $2015$-$04$ dump, which makes the graph somewhat overly-connected.
However, we find that counting the number of Wikilinks is still useful for measuring correlation between two entities.

\noindent $-$ {\em Transitive Categories} 
DBpedia entities are categorized according to the SKOS\footnote{\url{http://www.w3.org/2004/02/skos/}} vocabulary. 
Non-category entities are linked by predicate {\ttfamily subject} to the category entities they belong to.
One category is further linked to its super-category through predicate {\ttfamily broader}.
These categories and super-categories (around $1.2$ million in the $2015$-$04$ dump) can be found using transitive queries (see \cite{prud2008sparql} and Fig.~\ref{fig:flow_chart}), 
and are referred to as {\em transitive categories} here.

To answer the questions {\em IsThereAny}, {\em HowMany}, {\em IsTheA}, {\em LargestObj} and {\em AreAllThe},
we need to determine if there is a hyponymy relationship between two entities 
(\ie, one entity is conceptually a specific instance of another general entity).
This is done by checking if one entity is a transitive category of the other.
For question {\em CommProp}, we collect the transitive categories shared by two entities (see Fig.~\ref{fig:flow_chart}).
For questions {\em IsImgRelate} and {\em MostRelObj},
the correlation between a visual concept and the concept given in the question is measured based on checking the hyponymy relationship and 
counting the number of Wikilinks\footnote{We count the number of Wikilinks between two entities, 
but also consider paths through a third entity. See supplementary material for details.}.
Answering other templates (\eg, {\em FoodIngredient}) needs specific types of predicates extracted from Wikipedia infoboxes.
The complete list of queries for all templates is given in the supplementary material.

\noindent\textbf{Answer and Reason}
The last step is to generate answers according to the results of the queries.
Post-processing operations can be specified in Python within Quepy, and are needed for some questions, such as {\em IsImgRelate} and {\em MostRelObj} (see supplementary material for details).

Note that our system performs searches along the paths from 
visual concepts to KB concepts.
These paths can be used to give ``logical reasons'' as to how the answer is generated.
Especially for questions requiring external knowledge,
the predicates and entities on the path give a better understanding of how the relationships
are established between visual concepts and KB concepts.
Examples of answers and reasons can be seen in Fig.~\ref{fig:real_example} and \ref{fig:flow_chart}.

%% file: exp.tex
\vspace{-5pt}
\section{Experiments}
\noindent\textbf{Metrics}
Performance evaluation in VQA is complicated by the fact that two answers can have no words in common and yet both be perfectly correct.
Malinowski and Fritz~\cite{malinowski2014multi} used the Wu-Palmer similarity (WUPS) to measure the similarity between two words based on their common subsequence in the taxonomy tree. However, this evaluation metric restricts the answer to be a single word.  Antol~\etal~\cite{antol2015vqa} provided an evaluation metric for the open-answer task which records the percentage of answers in agreement with ground truth from several human subjects.
This evaluation metric requires around 10 ground truth answers for each question,
and only partly solves the problem (as indicated by the fact that even human performance is very low in some cases in~\cite{antol2015vqa}, such as `Why ...?' questions.).

In our case, the existing evaluation metrics are particularly unsuitable because most of the questions in our dataset are open-ended, especially for the ``\KBKnowl'' questions. In addition, there is no automated method for assessing the reasons provided by our system. The only alternative is a human-based approach.  Hence, we ask $5$ human subjects (examiners) to evaluate the results manually. In order to understand the generated answers better, we ask the examiners to give each answer or reason a correctness score as follows:
$1$: ``Totally wrong''; $2$: ``Slightly wrong''; $3$: ``Borderline''; $4$: ``OK''; $5$: ``Perfect''.
An answer or reason scored higher than
``Borderline'' %
is considered as ``right''; otherwise, it is considered as ``wrong''.
We perform this evaluation double-blind, \textit{i.e.} examiners are different from questions/answers providers and do not know the answers source. %
The protocol for measuring performance is described in the supplementary material.

\begin{table*}[t]
\centering
\small
\begin{tabular}{lccclccc}
\hline
\multicolumn{1}{c}{Question}  & \multicolumn{3}{c}{Accuracy($\%$)} &  & \multicolumn{3}{c}{Correctness (Avg.)} \\ \cline{2-4} \cline{6-8}
\multicolumn{1}{c}{Type}      & LSTM    & Ours    & Human    &  & LSTM       & Ours       & Human       \\ \hline
{\em IsThereAny}              & $64.9$    & $86.9$    & $93.6$     &  & $3.6$         & $4.5$        & $4.7$         \\
{\em IsImgRelate}             & $57.0$    & $82.2$    & $97.1$     &  & $3.3$         & $4.2$        & $4.9$         \\
{\em WhatIs}                  & $26.9$    & $66.9$    & $94.5$     &  & $2.1$         & $3.7$        & $4.8$         \\
{\em ImgScene}                & $30.4$    & $69.6$    & $85.9$     &  & $2.3$         & $3.8$        & $4.5$         \\
{\em ColorOf}                 & $14.6$    & $29.8$    & $93.2$     &  & $1.7$         & $2.5$        & $4.7$         \\
{\em HowMany}                 & $32.5$    & $56.1$    & $90.4$     &  & $2.3$         & $3.3$        & $4.6$         \\
{\em ObjAction}               & $19.7$    & $57.1$    & $90.5$     &  & $1.8$         & $3.5$        & $4.7$         \\
{\em IsSameThing}             & $54.9$    & $77.5$    & $91.5$     &  & $3.2$         & $4.2$        & $4.6$         \\
{\em MostRelObj}              & $32.1$    & $80.4$    & $92.9$     &  & $2.3$         & $4.2$        & $4.6$         \\
{\em ListObj}                 & $1.9$     & $63.0$    & $100$      &  & $1.1$         & $3.6$        & $4.8$         \\
{\em IsTheA}                  & $74.5$    & $80.4$    & $92.2$     &  & $3.9$         & $4.2$        & $4.7$         \\
{\em SportEquip}              & $2.1$     & $70.8$    & $79.2$     &  & $1.2$         & $3.9$        & $4.2$         \\
{\em AnimalClass}             & $0.0$     & $87.0$    & $95.7$     &  & $1.0$         & $4.5$        & $4.8$         \\
{\em LocIntro}                & $2.5$     & $67.5$    & $95.0$     &  & $1.1$         & $3.6$        & $4.8$         \\
{\em YearIntro}               & $0.0$     & $46.9$    & $93.8$     &  & $1.0$         & $2.9$        & $4.8$         \\
{\em FoodIngredient}          & $0.0$     & $58.1$    & $74.2$     &  & $1.0$         & $3.4$        & $4.3$         \\
{\em LargestObj}              & $0.0$     & $66.7$    & $96.3$     &  & $1.0$         & $3.8$        & $4.8$         \\
{\em AreAllThe}               & $29.6$    & $63.0$    & $81.5$     &  & $2.3$         & $3.7$        & $4.3$         \\
{\em CommProp}                & $0.0$     & $76.9$    & $76.9$     &  & $1.0$         & $4.1$        & $4.2$         \\
{\em AnimalRelative}          & $0.0$     & $88.2$    & $76.5$     &  & $1.1$         & $4.4$        & $4.1$         \\
{\em AnimalSame}              & $41.2$    & $70.6$    & $94.1$     &  & $2.6$         & $3.8$        & $4.8$         \\
{\em FirstIntro}              & $25.0$    & $25.0$    & $75.0$     &  & $2.0$         & $1.5$        & $4.1$         \\
{\em ListSameYear}            & $25.0$    & $75.0$    & $50.0$     &  & $1.8$         & $4.2$        & $3.0$         \\ \hline
{\em Overall}                 & $36.2$    & $69.6$    & $92.0$     &  & $2.5$         & $3.8$        & $4.7$         \\ \hline
\end{tabular}%
\caption{Human evaluation results of different methods for different question types. Accuracy is the percentage of correctly answered questions (\ie, correctness scored higher than ``Borderline'').
The average answer correctness ($\in [1,5]$, the higher the better) for each question type is also listed. %
We also evaluated human provided answers as a reference.}
\label{tab:result}
\end{table*}

\noindent\textbf{Evaluation}
We compare our \sexyname system with an approach (which we label LSTM) that encodes both CNN extracted features and questions with an encoder LSTM and generates answers with a decoder LSTM. Specifically, we use the second fully-connected layer ($4096$-d) of a pre-trained VGG model as the image features, and the LSTM is trained on the training set of COCO-VQA data~\cite{antol2015vqa}\footnote{
This baseline LSTM achieves $44.93\%$ accuracy on the COCO-VQA validation set, under its evaluation protocol. We note that training LSTM on another dataset provides it an unfair advantage. However, the current \KBName dataset is still relatively small and so does not support training large models. The presented dataset will be extended in near future.}.
The LSTM layer contains $256$ memory cells in each unit.
We also relate ``Human'' performance for reference.

Table~\ref{tab:result} provides the final evaluation results for different question types. Our system outperforms the LSTM on all question types with a final accuracy of $69.6\%$. Human performance is $92.0\%$. For question types particularly dependent on \KBKnowl, such as {\em AnimalClass}, {\em YearIntro}, {\em FoodIngredient}, {\em CommProp} and {\em AnimalRelative}, all LSTM-generated answers were marked as ``wrong'' by examiners. In contrast, our system performs very well on these questions. For example, we achieve $88.2\%$ on questions of type {\em AnimalRelative}, which is better than human performance. We also outperform humans on the question type {\em ListSameYear}, which requires  knowledge of the year a concept was introduced, and all things introduced in the same year. For the purely ``visual'' questions such as {\em WhatIs}, {\em HowMany} and {\em ColorOf}, there is still a gap between our proposed system and humans. However, this is mainly caused by the detection error of the object detectors, which is not the focus of this paper. According to the overall average correctness, we achieve $3.8$, which lies between ``Borderline'' and ``OK''. The LSTM scores only $2.5$ while Human achieves $4.7$.

\begin{figure}[tbp!]
\centering
\includegraphics[height=4.4cm]{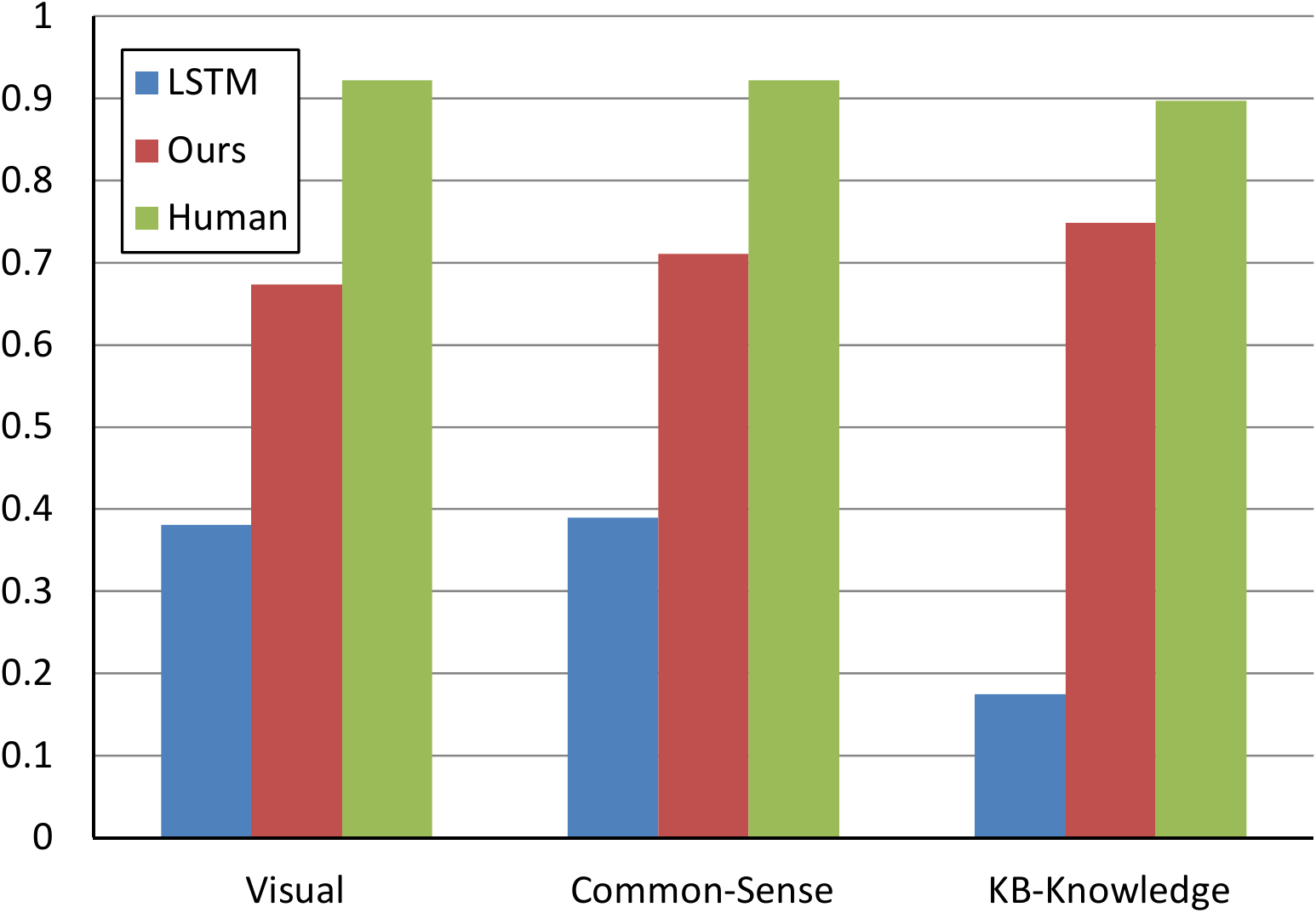}
\vspace{-6pt}
\caption{Accuracy of different methods for different knowledge levels. Humans perform almost equally over all three levels.
         LSTM performs worse for questions requiring higher-level knowledge, whereas \sexyname performs better.}
\label{chart:level_acc}
\vspace{-18.5pt}
\end{figure}

Fig.~\ref{chart:level_acc} relates the performance for ``Visual'', ``Common-sense'' and ``\KBKnowl'' questions. The overall trend is same as in Table~\ref{tab:result} --- \sexyname performs better than the LSTM method but not as well as humans. It is not surprising that humans perform almost equally for different knowledge levels, since we allow the human subjects to use Wikipedia to answer the ``\KBKnowl'' related questions. For the LSTM method, there is a significant decrease %
in performance as the dependency on external knowledge increases.
In contrast, \sexyname performs better as the level of external knowledge required increases.
{\em In summary, our system \sexyname performs better than LSTM at all three knowledge levels.
Furthermore, the performance gap between \sexyname and LSTM is more significant for questions requiring external knowledge. }

\begin{figure}[t]
\centering
\includegraphics[height=4.4cm]{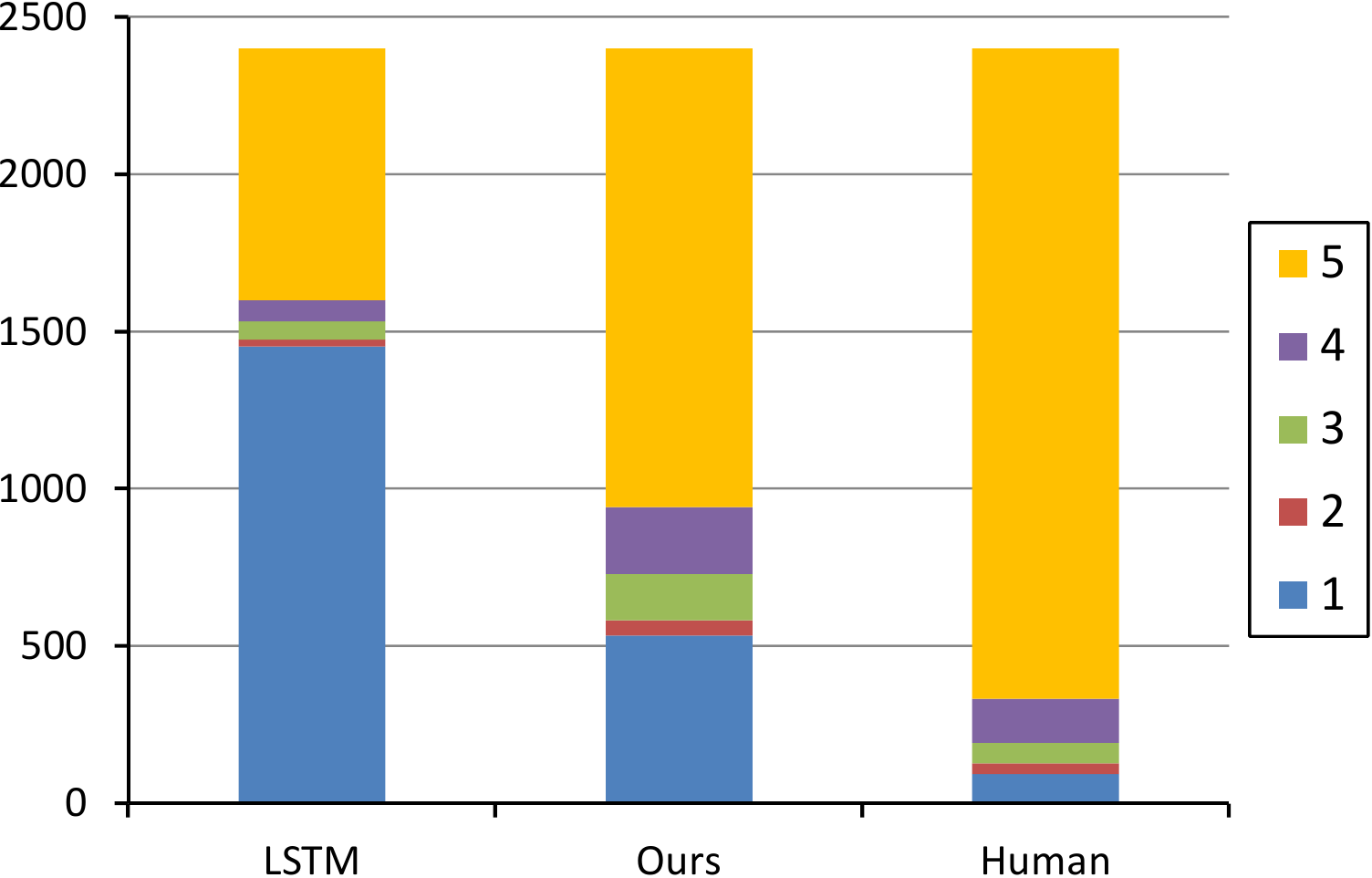}
\vspace{-3pt}
\caption{The number of answers that fall into the various correctness levels ($1$: ``Totally wrong''; $2$: ``Slightly wrong''; $3$: ``Borderline''; $4$: ``OK''; $5$: ``Perfect''),
         for different methods. }
\label{chart:correctness}
\vspace{-17pt}
\end{figure}

\begin{figure}[b]
\vspace{-15pt}
\centering
\includegraphics[width=0.9\linewidth]{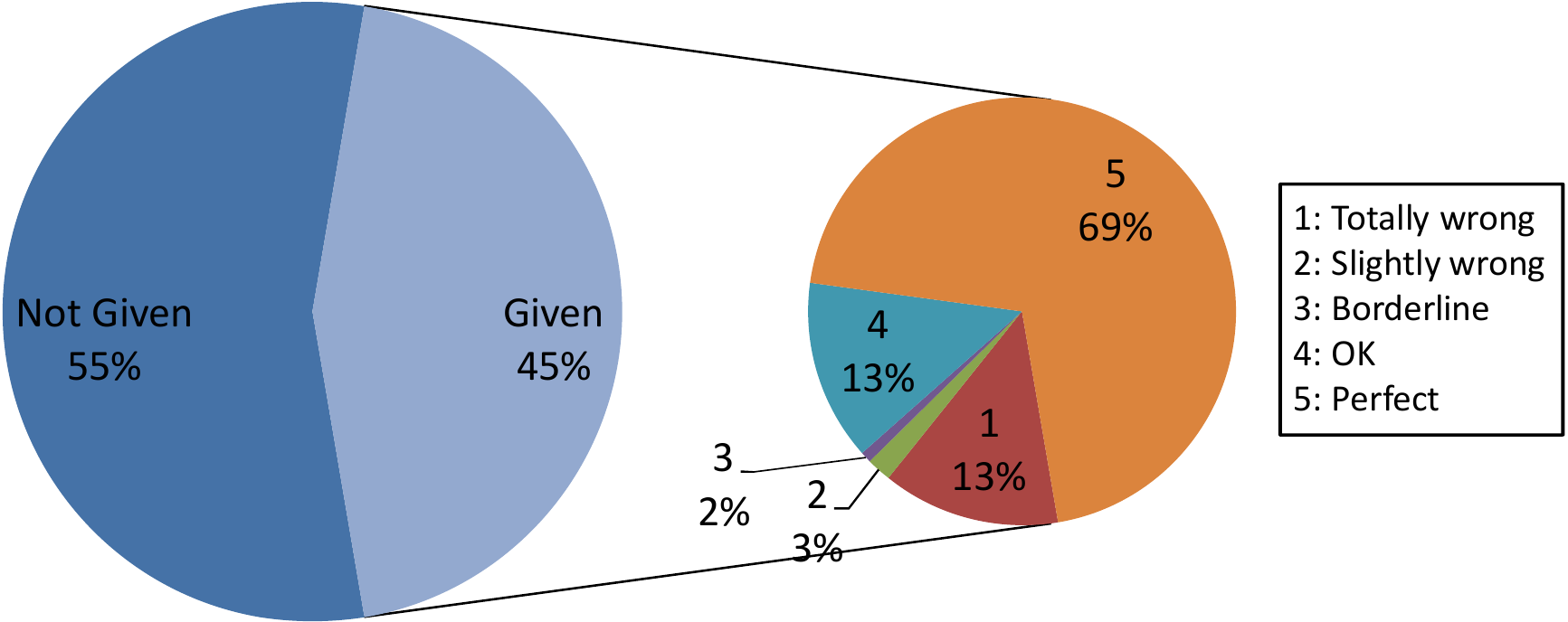}
\vspace{-1pt}
\caption{The correctness of the reasons provided by our \sexyname system.
         The left pie chart demonstrates that our system provides reasons for $45\%$ of the questions in the dataset.
         The right pie shows the distribution of correctness for the given reasons.}
\label{chart:correctness_reason}
\vspace{-3pt}
\end{figure}

\begin{figure*}[t!]
\centering
\includegraphics[width=1\linewidth]{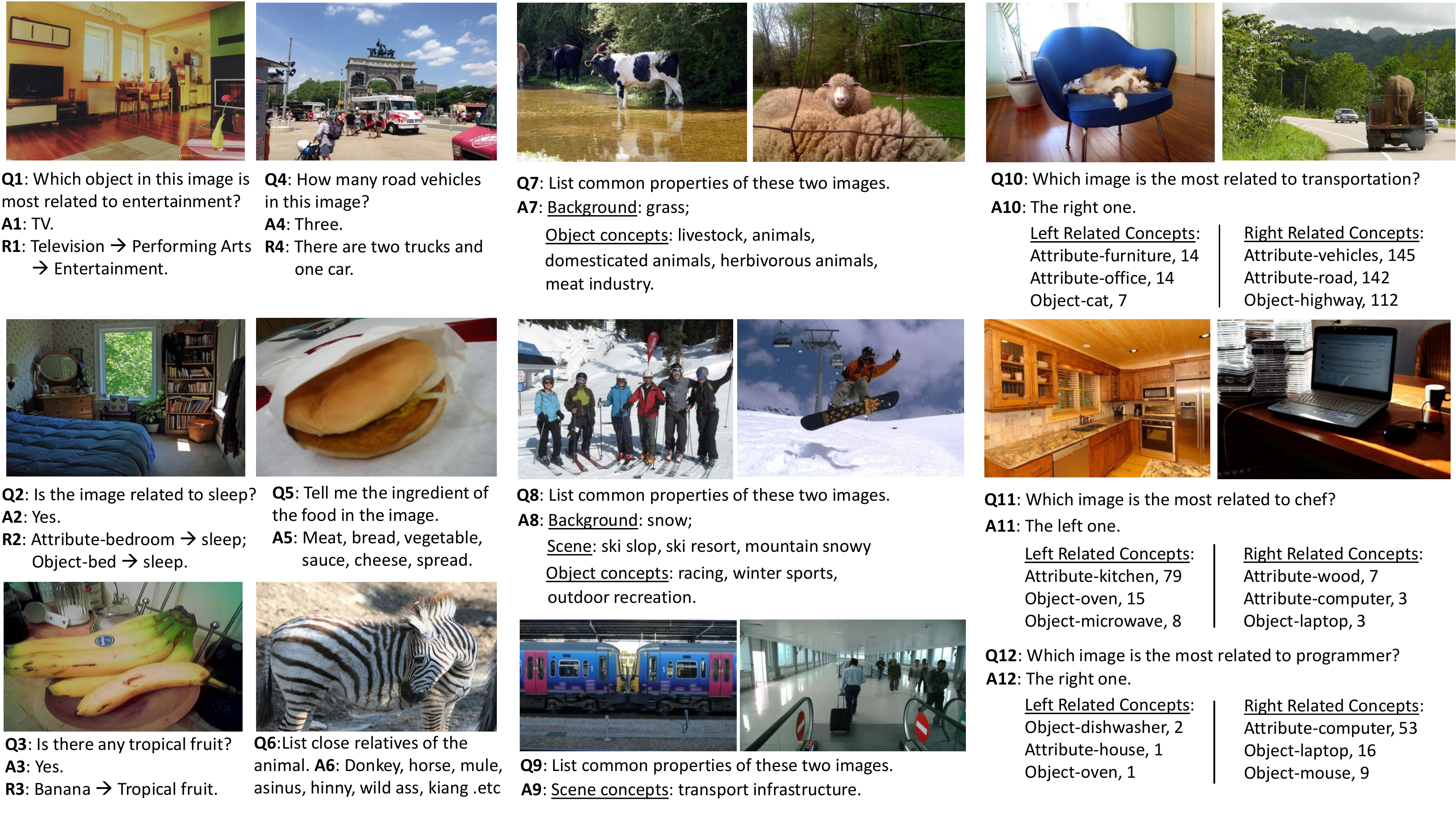}
\vspace{-19pt}
\caption{Examples of \KBName questions and the answers and reasons given by \sexyname.
Q$1$-Q$6$ are questions involving single images and Q$7$-Q$12$ are questions involving two images.
The numbers after visual concepts in Q$10$-Q$12$ are scores measuring the correlation between visual concepts and concepts given in questions (see supplementary material).}
\label{fig:final_results}
\vspace{-13pt}
\end{figure*}

Fig.~\ref{chart:correctness} indicates the numbers of answers that fall within the different correctness levels.
For the LSTM method, more than $50\%$ of the generated answers are grouped to level $1$, which is ``Totally wrong''.
For the human performance, only $10\%$ of the answers are not in level $5$.
\sexyname provides the largest portion (around $20\%$) of answers falling in levels $2$, $3$ or $4$.
From the view of the human examiners, the answers given by our \sexyname are ``softer'' than those of other methods.
Q$1$-Q$6$ of Fig.~\ref{fig:final_results} show some examples of the questions in \KBName dataset and the answers and reasons given by \sexyname.
More examples can be found in the supplementary material.

\noindent\textbf{Reason accuracy}
Fig.~\ref{chart:correctness_reason} gives an assessment of the quality of the reasons generated by our system.
Note that the LSTM cannot provide such information.
Since ``Visual'' questions can be answered by direct interrogation of pixels, we have not coded reasons into the question answering process for the corresponding templates (the reasons would be things like ``The corresponding pixels are brown'').
Fig.~\ref{chart:correctness_reason} relates the accuracy of the reasons given in the remaining $45\%$ of cases as measured by human examiners using the same protocol.
It shows that more than $80\%$ of reasons are marked as being correct (\ie, scored higher than $3$).
This is significant, as it shows that the method is using valid reasoning to draw its conculsions.
Examples of the generated reasons can be found in Fig.~\ref{fig:real_example}, Q$1$-Q$3$ in Fig.~\ref{fig:final_results} and in the supplementary material.

\noindent\textbf{Extending VQA forms}
Typically, a VQA problem involves one image and one natural language question (IMG$+$NLQ).
Here we extend  VQA to problems involving more images.
With this extension, we can ask more interesting questions and more clearly demonstrate the value of using a structured knowledge base.
In Fig.~\ref{fig:final_results}, we show two types of question involving two images and one natural language question (IMG1$+$IMG2$+$NLQ).
The first type of question (Q$7$-Q$9$) asks for the common properties between two whole images;
the second type (Q$10$-Q$12$) gives a concept and asks which image is the most related to this concept.

For the first question type,  \sexyname  obtains the answers
by searching all common transitive categories shared by the visual concepts extracted from the two query images.
For example, although the two images in Q$9$ are significantly different visually (even at the object level), and share no attributes in common,
their scene categories (railway station and airport) are linked to the same concept ``transport infrastructure'' in DBpedia.
For the second type, the correlation between each visual concept and the query concept is measured by a scoring function
(using the same strategy as is used for {\em IsImgRelate} and {\em MostRelObj} in Section~\ref{sec:answering_engines}),
and the correlation between an image and this concept is calculated by averaging the top three scores.
As we can see in Q$11$ and Q$12$, attributes ``kitchen'' and ``computer''
are most related to the concepts ``chef'' and ``programmer'' respectively,
so it is easy to judge that the answer for Q$11$ is the left image and the one for Q$12$ is the right.

The flexibility of Quepy, and the power of Python, make adding additional question types quite simple.  It would be straightforward to add question types requiring an image as an answer, for instance (IMG1+NLQ $\rightarrow$ IMGs).

%% file: appendix.tex
\makeatletter
\def\verbatim{\footnotesize\@verbatim \frenchspacing\@vobeyspaces \@xverbatim}
\makeatother

\section{Mapping $\langle${\em concept}$\rangle$ Phrases to KB-Entities with Redirections}
\label{sec:mapping}

Given a natural language phrase representing a real-world or abstract concept, we simply search its corresponding  KB-entities 
by matching the phrase to the name of KB-entities. 
The following SPARQL query is used in \sexyname to search the entities in DBpedia corresponding to ``Religion'':

\vspace{-3pt}
\begin{verbatim}
SELECT DISTINCT ?x WHERE {
  ?x label "Religion"@en. 
}
\end{verbatim}
\vspace{-3pt}

\noindent where the name of an entity is obtained through the predicate {\ttfamily label} (see Table~\ref{tab:predicates}).
The output of the above query is:

\vspace{-3pt}
\begin{verbatim}
http://dbpedia.org/resource/Category:Religion
http://dbpedia.org/resource/Religion
\end{verbatim}
\vspace{-3pt}

\noindent which shows that there are two KB entities matched to ``Religion''. 
The first one is a category entity and the second one is a non-category entity (see Section 3.2 of the main body).

Note that in natrual language, the same concept can be expressed in many ways, which can be caused by 
synonyms, capitalizations, punctuation, tenses, abbreviation and even misspellings.
For example, both ``Frisbee'' and ``Flying disc'' correspond to the disc-shaped gliding toy;
and the American technology company producing iPhone can be expressed as ``Apple Company'' and ``Apple Inc.''.

To handle this issue, 
DBpedia introduces a number of ``dummy'' entities that are only used to point to a ``concrete'' entity,
such that all different expressions of a concept are redirected to one concrete entity.

If the input phrase is the abbreviation ``Relig.'' rather than ``Religion'', the following SPARQL query can still locate the KB entity {\ttfamily KB:Religion}:%

\vspace{-3pt}
\begin{verbatim}
SELECT DISTINCT ?x1 WHERE {
  ?x0 label "Relig."@en. 
  ?x0 wikiPageRedirects ?x1. 
}
\end{verbatim}
\vspace{-3pt}

\noindent which outputs 

\vspace{-3pt}
\begin{verbatim}
http://dbpedia.org/resource/Religion
\end{verbatim}
\vspace{-3pt}

By using redirections, the vocabulary of concept phrases is enriched significantly.

\section{Query Generation and Post-processing}
\label{sec:query}

Given all the slot-phrases mapped to KB entities (see Section 3.2 in the main body), 
the next step is to generate queries specific to question templates, with post-processing steps for some templates.

The SPARQL queries used to answer specific questions are shown in the following.
There are two types of SPARQL queries (see \cite{prud2008sparql} for details): {\ttfamily ASK} queries check whether or not a query pattern has a solution, which returns $1$ if there is at least one solution and returns $0$ otherwise;
{\ttfamily SELECT} queries returns variables of all matched solutions.
In the following queries, the terms start with {\ttfamily ?} correspond to variables and others correspond to fixed entities or predicates.
Table~\ref{tab:predicates} shows the definition of the involved entities and predicates.

\noindent{\bf{\em WhatIs}}
The short abstract of the Wikipedia page corresponding to {\ttfamily KB:obj} is returned.

\vspace{-3pt}
\begin{verbatim}
SELECT ?desc WHERE {
  KB:obj comment ?desc. 
}
\end{verbatim}
\vspace{-3pt}

\noindent{\bf\em ColorOf}
The color of the mapped object is returned using the following query.

\vspace{-3pt}
\begin{verbatim}
SELECT DISTINCT ?obj_color { 
  Obj color ?obj_color. 
}
\end{verbatim}
\vspace{-3pt}

\noindent{\bf\em IsSameThing}
If the mapped two objects correspond to the same category name, the following query returns true; otherwise, it returns false.

\vspace{-3pt}
\begin{verbatim}
ASK { 
  KB:obj1 name ?obj_nm. 
  KB:obj2 name ?obj_nm. 
}
\end{verbatim}
\vspace{-3pt}

\noindent{\bf\em ListObj}
The category names of all objects contained in the questioned image are returned via the following query.

\vspace{-3pt}
\begin{verbatim}
SELECT DISTINCT ?obj_nm { 
  Img contain ?obj. 
  ?obj name ?obj_nm. 
}
\end{verbatim}
\vspace{-3pt}

\noindent{\bf\em ImgScene}
The scene information can be obtained from image attributes or scenes, using the following queries.
As attributes are trained with COCO data, they have higher priorities.

\vspace{-3pt}
\begin{verbatim}
SELECT DISTINCT ?att_nm { 
  Img  img-att ?att. 
  ?att supercat-name "scene". 
  ?att name ?att_nm.
}
\end{verbatim}
\vspace{-3pt}

\vspace{-3pt}
\begin{verbatim}
SELECT DISTINCT ?scn_nm { 
  Img  img-scn ?scn.
  ?scn name ?scn_nm.
}
\end{verbatim}
\vspace{-3pt}

\noindent{\bf\em ObjAction}
The following query returns the action attributes of an image.

\vspace{-3pt}
\begin{verbatim}
SELECT DISTINCT ?att_name { 
  Img  img-att ?att. 
  ?att supercat-name "action".
}
\end{verbatim}
\vspace{-3pt}

\noindent{\bf{\em IsThereAny}, {\em IsTheA}, {\em AreAllThe}, {\em LargestObj} and {\em HowMany}}
The following query checks whether or not {\ttfamily KB:concept} is a transitive category of {\ttfamily KB:obj},
which is used by the routines for the above five question types to determine 
if there is a hyponymy relationship between {\ttfamily KB:concept} and {\ttfamily KB:obj}.

\vspace{-3pt}
\begin{verbatim}
ASK { 
  KB:obj subject/broader?/broader? KB:concept. 
}
\end{verbatim}
\vspace{-3pt}

\noindent where a transitive closure is used (see \cite{prud2008sparql} for details) to find all transitive categories of {\ttfamily KB:obj} within three steps.
In this work, all transitive categories of an entity is limited to three steps, to avoid arriving at too general concepts.

For {\em IsThereAny}: true is returned if at least one object in the questioned image passes the above query (query returns $1$); otherwise, false is returned.

For {\em IsTheA}: if the identified object passes the query, return true; otherwise, return false.

For {\em AreAllThe}: return true, only if all identified objects pass the query.

For {\em LargestObj}: from all passed objects, return the one with the largest size.

For {\em HowMany}: return the number of passed objects.

\noindent{\bf{\em IsImgRelate} and {\em MostRelObj}}
For {\em MostRelObj}, the correlation of an object {\ttfamily KB:obj} and a concept {\ttfamily KB:concept}
is measured by the function 
\begin{equation}
\label{equ:scoring}
f = \alpha \cdot f_1 + f_2,
\end{equation}
where $f_1$ and $f_2$ are the respective outputs of the following two queries;
and $\alpha$ is the weight of $f_1$. 

\vspace{-3pt}
\begin{verbatim}
ASK { 
  { KB:obj WikiLink KB:concept } UNION
  { KB:concept WikiLink KB:obj } UNION
  { KB:obj subject/broader?/broader? KB:concept }.
}
SELECT COUNT(DISTINCT ?x) WHERE { 
  { KB:obj WikiLink ?x } UNION
  { ?x WikiLink KB:obj }.
  { KB:concept WikiLink ?x } UNION
  { ?x WikiLink KB:concept }.      
}
\end{verbatim}
\vspace{-3pt}

The first query returns $1$ if {\ttfamily KB:concept} is a transitive category of {\ttfamily KB:obj}, or
they are directly linked by predicate {\ttfamily WikiLink}.
The second query counts the number of indirect links via {\ttfamily WikiLink} and another KB-entity {\ttfamily ?x}.
Explicitly, the importance of $f_1$ is greater than $f_2$, so $\alpha$ is set to $50$ in the experiments.

For question {\em IsImgRelate}, the correlation of the concept and each object/attribute/scene of the questioned image is measured by 
the same scoring function \eqref{equ:scoring}.
If any score is larger than $50$, the image is considered as being related to the concept.

This scoring function is used for
question ``Which image is the most related to $\langle${\em concept}$\rangle$?'' as well.

\noindent{\bf{\em CommProp}}
The transitive categories shared by both {\ttfamily KB:obj1} and {\ttfamily KB:obj2} are returned through:

\vspace{-3pt}
\begin{verbatim}
SELECT DISTINCT ?cat WHERE { 
  KB:obj1 subject/broader?/broader? ?cat.
  KB:obj2 subject/broader?/broader? ?cat.
}
\end{verbatim}
\vspace{-3pt}

\noindent Note that {\ttfamily KB:obj2} can be replaced by {\ttfamily KB:concept} based on the asked question.

For question ``List common properties of these two images.'',
the common transitive categories are obtained by comparing the objects/attributes/scenes of two images.

\noindent{\bf{\em SportEquip}}
In DBpedia, the equipments for a specific sport share a common KB-category.
This KB-category has the label ``$\langle$sport-name$\rangle$ equipment'', where $\langle$sport-name$\rangle$ is the name of a sport (such as ``Tennis'', ``Skiing'').
The sport name can be obtained from image attributes.
The following query returns all tennis equipments:

\vspace{-3pt}
\begin{verbatim}
SELECT DISTINCT ?equip WHERE {
  ?equip subject ?cat. 
  ?cat   label ?cat_nm. 
  FILTER regex(?cat_nm, "Tennis equipment").
}
\end{verbatim}
\vspace{-3pt}

\noindent which uses a regular expression (regex) to parse the label of KB-categories.

If the sport information is not contained in attributes.
We can answer this question by the following query, 
which returns all equipments sharing the same ``$\langle$sport-name$\rangle$ equipment''-category with a detected object in the image.

\vspace{-3pt}
\begin{verbatim}
SELECT DISTINCT ?equip WHERE {
  Img    contain ?obj.  
  ?obj    subject ?cat. 
  ?equip subject ?cat. 
  ?cat   broader/broader? KB:Cat-Sports_equipment.
}
\end{verbatim}
\vspace{-3pt}

\noindent where the entity {\ttfamily KB:Cat-Sports\_equipment}\footnote{The URI is \url{http://dbpedia.org/resource/Category:Sports_equipment}} 
is a super-category of all ``$\langle$sport-name$\rangle$ equipment''-categories.

\begin{figure*}[t!]
\centering
\includegraphics[width=0.85\linewidth]{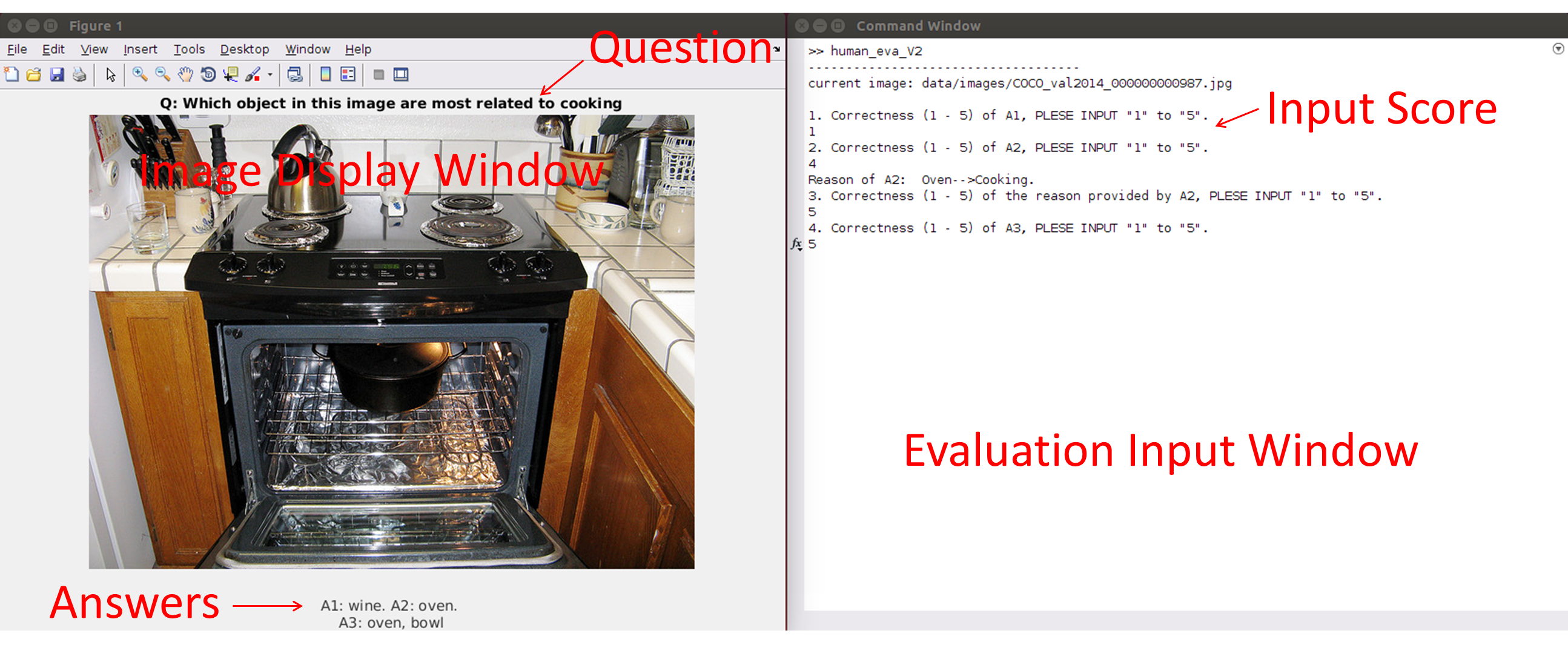}
\caption{Our evaluation tool interface, written in Matlab. The left side is the image, question and answers displaying side. The right side is the user input side. Answers are labelled as A1, A2 and A3, so the examiner has no clue about the source of the answers. During the evaluation, users (examiners) are only required to input a score from $1$ to $5$ for each evaluation item.}
\label{fig:eva_tool}
\end{figure*}

\noindent{\bf{\em LocIntro}}
The place where a KB-entity is invented is recorded by a KB-category with label ``$\langle$country$\rangle$ inventions''.

\vspace{-3pt}
\begin{verbatim}
SELECT DISTINCT ?cat_nm WHERE {
  KB:obj subject ?cat. 
  ?cat   label ?cat_nm.
  FILTER regex(?cat_nm,"^[a-z|A-z]+ inventions$").
}
\end{verbatim}
\vspace{-3pt}

\noindent{\bf{\em YearIntro}, {\em FirstIntro} and {\em ListSameYear}}
The introduction year of a KB-entity is recorded by a KB-category with label ``$\langle$year$\rangle$ introductions''.
The following query returns the introduction year of {\ttfamily KB:obj}, which is used for question type {\em YearIntro}:

\vspace{-3pt}
\begin{verbatim}
SELECT DISTINCT ?cat_nm WHERE {
  KB:obj subject ?cat. 
  ?cat   label ?cat_nm.
  FILTER regex(?cat_nm,"^[0-9]+ introductions$").
}
\end{verbatim}
\vspace{-3pt}

For {\em FirstIntro}, the introduction years of two entities are compared.

For {\em ListSameYear}, the following query returns all things introduced in the same year of {\ttfamily KB:obj}:

\vspace{-3pt}
\begin{verbatim}
SELECT DISTINCT ?thing_nm WHERE {
  ?thing subject ?cat.
  ?thing label ?thing_nm.
  KB:obj subject ?cat. 
  ?cat   label ?cat_nm.
  FILTER regex(?cat_nm,"^[0-9]+ introductions$").
}
\end{verbatim}
\vspace{-3pt}

\noindent{\bf\em FoodIngredient}
With predicate {\ttfamily ingredient}, the following query returns the ingredients of {\ttfamily KB:food}.

\vspace{-3pt}
\begin{verbatim}
SELECT DISTINCT ?Ingrd_nm WHERE { 
  KB:food ingredient ?Ingrd.
  ?Ingrd  label      ?Ingrd_nm. 
}
\end{verbatim}
\vspace{-3pt}

\noindent{\bf{\em AnimalClass}, {\em AnimalRelative} and {\em AnimalSame}}
With predicate {\ttfamily taxonomy} (see Table~\ref{tab:predicates}), the following queries are used for 
the above three templates respectively.

\vspace{-3pt}
\begin{verbatim}
SELECT DISTINCT ?class_nm WHERE { 
  KB:animal taxonomy ?class.
  ?class    label    ?class_nm. 
}
SELECT DISTINCT ?relative_nm WHERE { 
  KB:animal taxonomy ?class.
  ?relative taxonomy ?class.
  ?relative label    ?relative_nm. 
}
ASK { 
  KB:animal1 taxonomy ?class.
  KB:animal2 taxonomy ?class.
}
\end{verbatim}
\vspace{-3pt}

\begin{table*}[t!]
\centering
\resizebox{0.95\linewidth}{!}{\footnotesize
\centering
\begin{tabular}{llp{11cm}}
\hline
Term                    & Defined by & Description \\ \hline\hline
& &\\ [-2ex]
{\bf Entities} & & \\
& &\\ [-2ex]
{\ttfamily Img}         & \sexyname & The \sexyname entity corresponding to the questioned image. \\
& &\\ [-2ex]
{\ttfamily Obj} 	& \sexyname & The \sexyname entity mapped to slot $\langle${\em obj}$\rangle$. \\
& &\\ [-2ex]
{\ttfamily KB:obj}	& DBpedia  & The KB-entity mapped to slot $\langle${\em obj}$\rangle$. \\
& &\\ [-2ex]
{\ttfamily KB:concept}	& DBpedia  & The KB-entity mapped to slot $\langle${\em concept}$\rangle$. \\
& &\\ [-2ex]
{\ttfamily KB:food}     & DBpedia  & The KB-entity mapped to slot $\langle${\em food}$\rangle$. \\
& &\\ [-2ex]
{\ttfamily KB:animal}   & DBpedia  & The KB-entity mapped to slot $\langle${\em animal}$\rangle$. \\
& &\\ [-2ex]
\hline
& &\\ [-2ex]
{\bf Predicates} & & \\
& &\\ [-2ex]
{\ttfamily img-att}       & \sexyname & Linking an image to one of its attribute categories.\\
& &\\ [-2ex]
{\ttfamily img-scn}       & \sexyname & Linking an image to one of its scene categories.\\
& &\\ [-2ex]
{\ttfamily contain}       & \sexyname & Linking an image to one of the objects it contains.\\
& &\\ [-2ex]
{\ttfamily name}          & \sexyname & Linking an object/scene/attribute category to its name (string).\\ 
& &\\ [-2ex]
{\ttfamily color}         & \sexyname & Linking an object to its color.\\
& &\\ [-2ex]
{\ttfamily size}          & \sexyname & Linking an object to its size.\\
& &\\ [-2ex]
{\ttfamily supercat-name} & \sexyname & Linking an object/scene/attribute category to its super-category name (string).\\
& &\\ [-2ex]
{\ttfamily label}         & DBpedia   & Linking a KB-entity to its name (string). \\
                                  & &   Its URI is \url{http://www.w3.org/2000/01/rdf-schema\#label}. \\
& &\\ [-2ex]
{\ttfamily comment}       & DBpedia   & Linking a KB-entity to its short description. \\
                                  & &   Its URI is \url{http://www.w3.org/2000/01/rdf-schema\#comment}. \\
& &\\ [-2ex]
{\ttfamily wikiPageRedirects} & DBpedia   
                                      & Linking a ``dummy'' KB-entity to the ``concrete'' one describing the same concept. \\
                                  & &  Its URI is \url{http://dbpedia.org/ontology/wikiPageRedirects}\\
& &\\ [-2ex]
{\ttfamily subject}       & DBpedia   & Linking a non-category KB-entity to its categories. \\
                                  & &   Its URI is \url{http://purl.org/dc/terms/subject}. \\
& &\\ [-2ex]
{\ttfamily broader}       & DBpedia   & Linking a category KB-entity to its super-categories. \\
                                  & &   Its URI is \url{http://www.w3.org/2004/02/skos/core\#broader}.\\
& &\\ [-2ex]
{\ttfamily Wikilink}      & DBpedia   & Linking two correlated non-category KB-entities, which is extracted from the internal links bettween Wikipedia articles. \\
                                  & &   Its URI is \url{http://dbpedia.org/ontology/wikiPageWikiLink}. \\
& &\\ [-2ex]
{\ttfamily ingredient}    & DBpedia   & Linking a food KB-entity to its ingredient, which is extracted from infoboxes of Wikipedia. \\
                                  & &   Its URI is \url{http://dbpedia.org/ontology/ingredient}. \\
& &\\ [-2ex]
{\ttfamily taxonomy}      & DBpedia   & Linking an animal KB-entity to its taxonomy class (can be kingdom, phylum, class, order, family or genus), which is extracted from infoboxes of Wikipedia. \\
                                  & &   For phylum, the URI is \url{http://dbpedia.org/ontology/phylum}.\\
& &\\ [-2ex]
\hline
\end{tabular} }
\vspace{-5pt}
\caption{The RDF entities and predicates used in \sexyname, some of which are defined by \sexyname and others are originally defined by DBpedia.}
\vspace{-5pt}
\label{tab:predicates}
\end{table*}

\section{Evaluation Protocol}
\label{sec:evaluation}

\input{eva_protocol.tex}

\section{Visual Concepts}
\label{sec:visual}

Three types of visual concepts are detected in \sexyname, which are object, attributes and scenes.
The scene classes are defined the same as in the MIT Places$205$ dataset~\cite{zhou2014learning}.
The object classes are obtained by merging the $80$ classes in MS COCO object dataset~\cite{lin2014microsoft} 
and $200$ classes in ImageNet object dataset~\cite{deng2009imagenet},
which are shown in Table~\ref{tab:object}.
The vocabulary of attributes trained in \cite{qi2015caption} is demonstrated in Table~\ref{tab:att}.

\begin{table*}[tbp!]
\centering
\footnotesize{
\begin{tabular}{llp{12cm}}
\hline
Super-category & Number & Attribute categories                                                                                                                                                                                                                                                                                                                                                                                                                                                                                                                                                                                                                                                                                                                                                                                                                                                                                                                                                      \\ \hline\hline
Action  & $24$     & playing, sitting, standing, swinging, catching, cutting, dining, driving, eating, flying, hitting, jumping, laying, racing, reads, swimming, running, sleeping, smiling, taking, talking, walking, wearing,  wedding 
\\ \hline
Sport  &  $4$ &      surfing, tennis, baseball, skateboard                                                                                                                                                                                                                                                                                                                                                                                                                                                                                                                                                                                                                                                                                                                                \\ \hline
Scene          & $24$     & mountain, road, snow, airport, bathroom, beach, bedroom, city, court, forest, hill, island, kitchen, lake, market, ocean, office, park, river, room, sea, sky, restaurant, field zoo                                                                                                                                                                                                                                                                                                                                                                                                                                                                                                                                                                                                                                                                                                                                                                                                \\ \hline
Object         & $95$     & children, bottle, computer, drink, glass, monitor, tree, wood, basket, bathtub, beer, blanket, box, bread, bridge, buildings, cabinets, camera, candles, cheese, chicken, chocolate, church, clouds, coat, coffee, decker, desk, dishes, door, face, fence, fish, flag, flowers, foods, fruits, furniture, grass, hair, hands, head, helmet, hotdog, house, ice, jacket, kitten, lettuce, lights, luggage, meat, metal, mouth, onions, palm, pants, papers, pen, pillows, plants, plates, players, police, potatoes, racquet, railing, rain, rocks, salad, sand, seat, shelf, ship, shirt, shorts, shower, sofa, station, stone, suit, sunglasses, toddler, tomatoes, towel, tower, toys, tracks, vegetables, vehicles, wall, water, wii, windows, wine \\ \hline
\end{tabular}
}
\caption{The $147$ image atributes used as visual concepts in \sexyname system. }
\label{tab:att}
\end{table*}

\begin{table*}[tbp!]
\centering
\footnotesize{
\begin{tabular}{llp{12cm}}
\hline
Super-category & Number & Object categories                                                                                                                                                                                                                                                                                                                                                                                           \\ \hline\hline
person         & $1$      & person                                                                                                                                                                                                                                                                                                                                                                                                      \\ \hline
vehicle        & $12$     & bicycle, car, motorcycle, airplane, bus, train, truck, boat, cart, snowmobile, snowplow, unicycle                                                                                                                                                                                                                                                                                                           \\ \hline
outdoor        & $5$      & traffic light, fire hydrant, stop sign, parking meter, bench                                                                                                                                                                                                                                                                                                                                                \\ \hline
animal         & $48$     & cat, dog, horse, sheep, cow, elephant, bear, zebra, giraffe, ant, antelope, armadillo, bee, butterfly, camel, centipede,dragonfly, fox, frog, giant panda, goldfish, hamster, hippopotamus, isopod, jellyfish, koala bear, ladybug, lion, lizard, lobster,monkey, otter, porcupine, rabbit, ray, red panda, scorpion, seal, skunk, snail, snake, squirrel, starfish, swine, tick, tiger, turtle,whale \\ \hline
accessory      & $18$    & backpack, umbrella, handbag, tie, suitcase, band aid, bathing cap, crutch, diaper, face powder, hat with awide brim, helmet, maillot, miniskirt, neck brace, plastic bag, stethoscope, swimming trunks                                                                                                                                                                      \\ \hline
sports         & $22$     & frisbee, skis, snowboard, kite, baseball bat, baseball glove, skateboard, surfboard, tennis racket, balance beam, baseball,basketball, croquet ball, golf ball, golfcart, horizontal bar, punching bag, racket, rugby ball,soccer ball, tennis ball, volleyball                                                                                                             \\ \hline
kitchen        & $20$     & bottle, wine glass, cup, fork, knife, spoon, bowl, beaker, can opener, cocktail shaker, corkscrew, frying pan, ladle, milk can,pitcher, plate rack, salt or pepper shaker, spatula, strainer, water bottle                                                                                                                                                                                                  \\ \hline
food           & $27$     & banana, apple, sandwich, orange, broccoli, carrot, hot dog, pizza, donut, cake, artichoke, bagel, bell pepper, burrito, cream,cucumber, fig, guacamole, hamburger, head cabbage, lemon, mushroom, pineapple, pomegranate, popsicle, pretzel,strawberry                                                                                                                                                      \\ \hline
furniture      & $8$      & chair, couch, potted plant, bed, dining table, toilet, baby bed, filing cabinet                                                                                                                                                                                                                                                                                                                             \\ \hline
electronic     & $7$      & tv, laptop, mouse, remote, keyboard, cell phone, iPod                                                                                                                                                                                                                                                                                                                                                       \\ \hline
appliance      & $14$     & microwave, oven, toaster, sink, refrigerator, coffee maker, dishwasher, electric fan, printer, stove, tape player, vacuum,waffle iron, washer                                                                                                                                                                                                                                                               \\ \hline
indoor         & $18$     & clock, vase, scissors, teddy bear, hair drier, toothbrush, binder, bookshelf, digital clock, hair spray, lamp, lipstick, pencil box, pencil sharpener, perfume, rubber eraser, ruler, soap dispenser                                                                                                                                                                                             \\ \hline
music          & $16$     & accordion, banjo, cello, chime, drum, flute, french horn, guitar, harmonica, harp, oboe, piano, saxophone,trombone, trumpet, violin                                                                                                                                                                                                                                                                 \\ \hline
tool           & $8$      & axe, bow, chain saw, hammer, power drill, screwdriver, stretcher, syringe                                                                                                                                                                                                                                                                                                                                   \\ \hline
\end{tabular}
}
\caption{The $224$ object classes used as visual concepts in \sexyname system.}
\label{tab:object}
\end{table*}

\section{Examples}
\label{sec:examples}
\begin{figure*}[tbp!]
\begin{center}
   \includegraphics[width=0.8\linewidth]{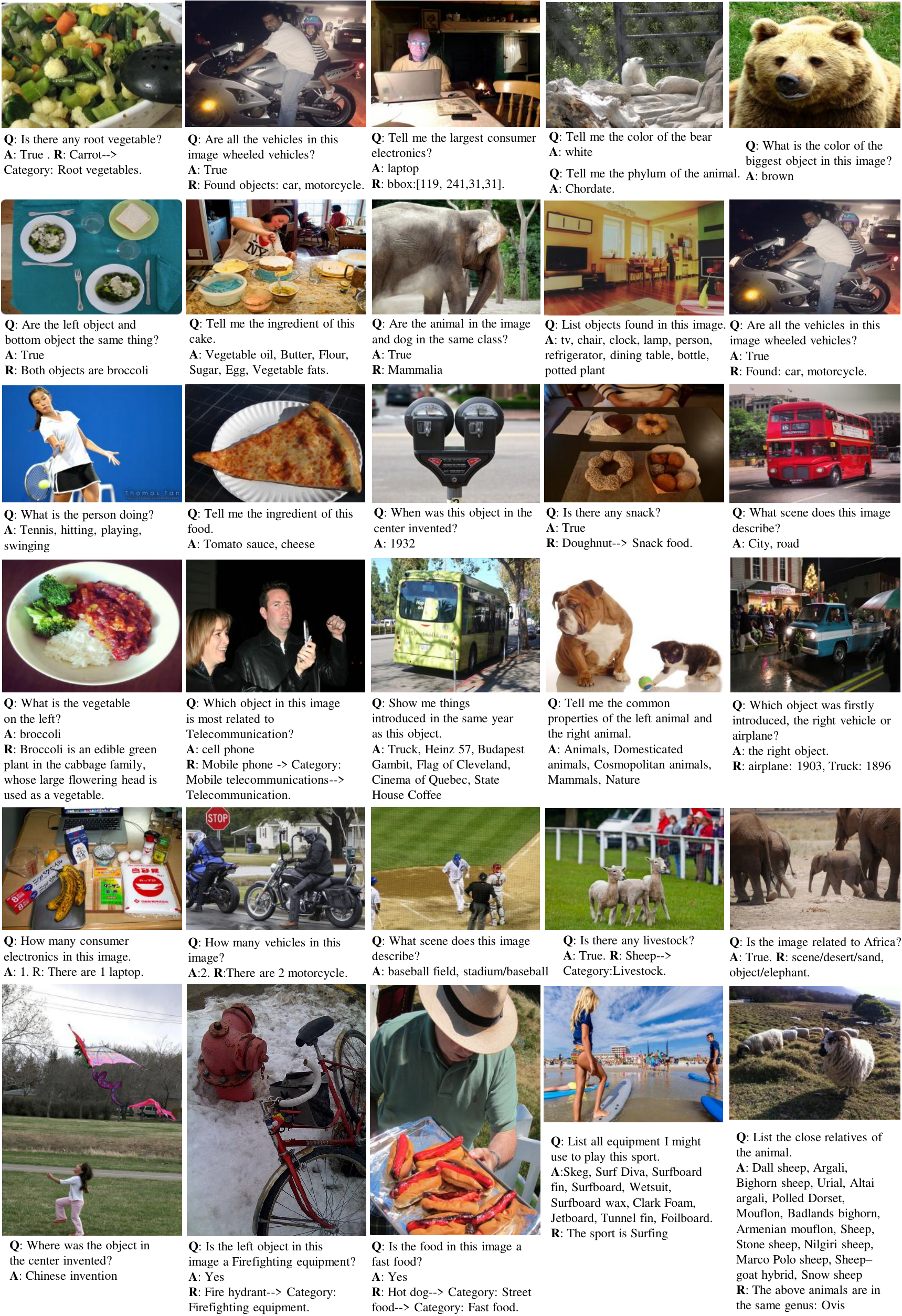}
\end{center}
\vspace{-15pt}
   \caption{Examples of KB-VQA questions and the answers generated by \sexyname. {\bf Q}: questions; {\bf A}: answers; {\bf R}: reasons.
            }
\vspace{-15pt}
\label{fig:example1}
\end{figure*}

Examples of KB-VQA questions and the answers generated by \sexyname are shown in Fig.~\ref{fig:example1}.

%
%
%
%

%

%% file: eva_protocol.tex
A user-friendly interface (in Matlab) is developed for the human subject (examiner) evaluation. During the time of evaluation, a question, an image and three answers are popped up. Those answers are generated by LSTM, our system and the other human subject, respectively (see Figure~\ref{fig:eva_tool}). Our evaluation process is double-blind, that means, examiners are different from those questioners (who ask questions). Moreover, the answer sources (\ie, which algorithm generated the answer) are not provided to the examiner.

We ask the examiners to give a correctness score ($1$-$5$) for each answer as following:

\begin{enumerate}
 \item[1:] Totally wrong (if the examiner thinks the candidate answer is largely different from it suppose to be or even wrong answer type).
 \item[2:] Slightly wrong (if the examiner thinks the candidate answer is wrong, but close to right. For example, the groundtruth answer is ``pink'' while the generated answer is ``red'').
 \item[3:] Borderline (applicable when the groundtruth answer is a list. if the examiner thinks the candidate answers only hit little amount of the answers in the list).
 \item[4:] OK (applicable when the groundtruth answer is a list. if the examiner thinks the candidate answers hit most of the answers in the list).
 \item[5:] Perfect (if the examiner thinks the candidate answer is perfectly correct).
\end{enumerate}
Finally, the overall accuracy for a given question type is calculated by $\frac {\text{\# of answers that correctness}>3}{\text{Total \# of questions}}$. The same rule is applied for the ``logical reason'' correctness evaluation.